\pdfoutput=1

\documentclass{article}

\usepackage[preprint]{acl}

\usepackage{times}
\usepackage{latexsym}
\usepackage{adjustbox}
\usepackage{multirow}

\usepackage[T1]{fontenc}

\usepackage[utf8]{inputenc}

\usepackage{microtype}

\usepackage{inconsolata}

\usepackage{graphicx}
\usepackage{booktabs}
\usepackage{mdframed}

\usepackage{enumitem}
\usepackage{subcaption}

\usepackage{colortbl} 
\definecolor{mypink}{RGB}{255,221,168}
\definecolor{mycolor}{RGB}{255, 182, 193}

\usepackage{url}
\usepackage{multirow}
\usepackage{booktabs,subcaption,amsfonts,dcolumn}
\usepackage{xspace}
\usepackage{amsmath}
\usepackage{algorithm}
\usepackage{algpseudocode}
\newcommand{\thickhline}{\noalign{\hrule height 1.0pt}}

\newcommand{\ours}{DPRefine\xspace}

\title{Differentially Private Learning Needs Better Model Initialization and Self-Distillation}

\author{
\normalfont
\textbf{Ivoline C. Ngong}\textsuperscript{1} \,
\textbf{Joseph P. Near}\textsuperscript{1} \,
\textbf{Niloofar Mireshghallah}\textsuperscript{2}\\
\vspace{5pt} \\ 
\textsuperscript{1}University of Vermont \,
\textsuperscript{2}University of  Washington \\
\texttt{kngongiv@uvm.edu,jnear@uvm.edu,niloofar@cs.washington.edu}
}

\renewcommand{\paragraph}[1]{\vspace*{1pt}\noindent\textbf{#1}}
\begin{document}
\maketitle
\begin{abstract}
Differentially private SGD (DPSGD) enables privacy-preserving training of language models, but often reduces utility, diversity, and linguistic quality. We introduce DPRefine, a three-phase method that initializes a model using data synthesis from a small pre-trained LM with rigorous filtering, applies DP finetuning on private data, and performs self-distillation to refine outputs. This approach significantly outperforms vanilla DPSGD, with AlpacaEval preferring DPRefine's generations in 78.4\% of cases across all datasets. Our analysis reveals that DPRefine reduces linguistic errors in generated text by 84.0\%, mitigating grammar and spelling errors, commonly associated with DPSGD. It also reduces inconsistencies of non-private models, such as hallucinated details and misattributed quotes. We find that small models like GPT-2 can be effective for initialization and distillation, highlighting their potential in enabling scalable and efficient deployment of privacy-preserving language.
\end{abstract}

\section{Introduction}
Training machine learning models on private data offers significant potential for enhancing performance on domain-specific tasks, particularly in natural language processing \cite{li2021large,yu2021differentially,liu2023differentially,cummings2023advancing}. However, the use of sensitive information raises critical privacy concerns. Differentially Private Stochastic Gradient Descent (DPSGD) has emerged as a prominent technique to bound information leakage during model training by clipping gradients and adding calibrated noise during optimization \cite{abadi2016deep}. While DPSGD provides strong privacy guarantees, the introduced noise and gradient modifications lead to significant challenges: decreased model utility \cite{yu2021differentially,ponomareva2022training}, less diverse text generation due to distribution smoothing \cite{bagdasaryan2019differential,mireshghallah2022privacy}, and notably, as our analysis reveals, increased linguistic errors in generated text (Table~\ref{tbl:hallucination_examples}).

A common scenario in practice involves domain-specific tasks with limited private labeled data, where organizations aim to leverage existing pre-trained language models while preserving privacy. Simply applying DPSGD to fine-tune these models on private data often yields poor results, particularly when the private dataset is small~\cite{tramer2022considerations,mireshghallah2021privacy}. Recent work has shown that leveraging better hand-crafted features \cite{tramer2020differentially} or features from large pre-trained language models \cite{li2022does,li2021large} can improve the privacy-utility trade-off in differentially private learning. However, these approaches have limitations: smaller pre-trained models offer limited benefits,  and fine-tuning larger models on private data may be infeasible due to proprietary concerns or infrastructure limitations. This raises a critical question: \textit{Can we develop small, domain-specific language models that achieve high performance without requiring large private datasets or large, pre-trained models?}

\begin{figure*}
    \centering
    \includegraphics[width=\textwidth]{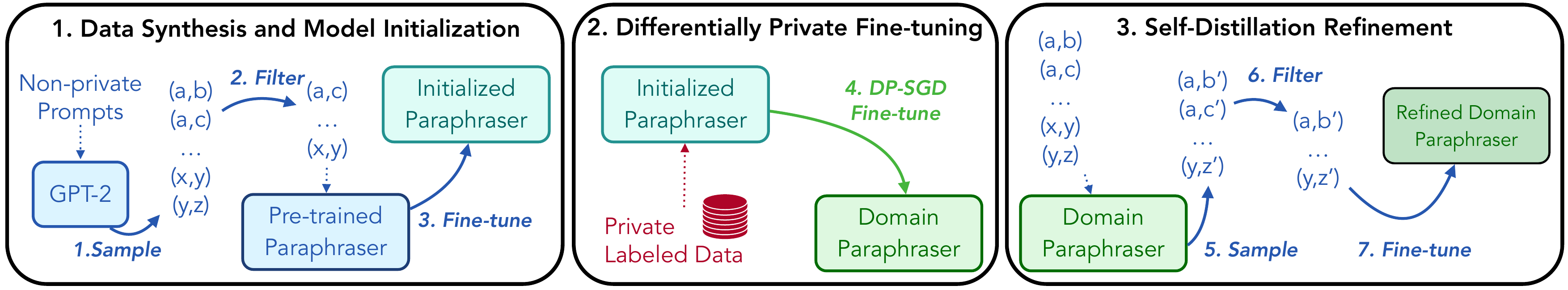}
\caption{Overview of DPRefine's three-phase approach: Phase 1: Data Synthesis and Model Initialization generates synthetic training pairs using GPT-2, applies quality filtering, and performs initial fine-tuning on a pre-trained  T5 (encoder-decoder model) to create an initialized paraphraser/summarize, all without accessing private data. Phase 2: Differentially Private Fine-tuning applies DP-SGD on private labeled data to create a privacy-preserving domain paraphraser/summarizer. Phase 3: Self-Distillation Refinement uses the DP model to generate new training pairs, applies filtering, and performs final fine-tuning to produce a refined domain paraphraser/summarizer.}
    \label{fig:conceptual_understanding_tool}
\end{figure*}

In this paper we introduce DPRefine, building on the intuition that model performance under differential privacy heavily depends on model initialization and feature representation~\cite{tramer2020differentially,li2022does}. Additionally, recent advances in NLP have demonstrated the effectiveness of data synthesis~\cite{liu2024best,flemings2024differentially}, filtering and self-distillation in improving the performance of smaller models\cite{jung2023impossible,bansal2024smaller}.
Our method has three phases, depicted in Figure~\ref{fig:conceptual_understanding_tool}: In the first phase we create a strong initialization by generating high-quality synthetic data using a small pre-trained language model (e.g., GPT-2) with rigorous filtering - importantly, this phase operates entirely independently of any private data. In the second phase we apply DPSGD to fine-tune this initialized model on the private labeled data, representing the only stage where private data is accessed. 

Finally, we apply self-distillation where the DP model generates new training data for further refinement, again without accessing the original private data.
We evaluate DPRefine on three datasets across two domain-specific tasks: summarization, XSum~\cite{narayan2018don} and PubMed~\cite{pubmed}, and paraphrasing, MRPC~\cite{dolan-brockett-2005-automatically}. Our experiments demonstrate that DPRefine significantly outperforms vanilla DPSGD, with AlpacaEval~\cite{li2023alpacaeval} preferring DPRefine's generations in 78.38\% of cases across all datasets and metrics.

Beyond standard metrics, we conduct a fine-grained manual error analysis of the generated text, constructing a taxonomy of both linguistic errors (e.g., grammar and spelling mistakes) and non-linguistic errors, which we refer to throughout the paper as inconsistencies and hallucinations (e.g., misattributed quotes or ungrounded contextual details), as shown in Table~\ref{tbl:hallucination_examples}. Our analysis reveals that DPRefine reduces linguistic errors by 84.0\% compared to vanilla DPSGD while also mitigating inconsistencies present in non-private models.
Our implementation of DPRefine is available as open source.\footnote{\url{https://github.com/uvm-plaid/private_llm_generation}}




\section{Preliminaries}
\paragraph{Differential Privacy: }
Differential privacy (DP)~\cite{dwork2006calibrating, dwork2014algorithmic} is a formal privacy definition that provides strong guarantees by limiting the influence any single data point can have on the output of an algorithm. Formally, a randomized mechanism $\mathcal{M}$ satisfies $(\epsilon, \delta)$-DP if for any 2 neighboring datasets $D, D' \in \mathcal{D}$ that differ in exactly one data sample, and for all sets of outcomes $S$, $\Pr[\mathcal{M}(D) \in S] \leq e^\epsilon \Pr[\mathcal{M}(D') \in S] + \delta$.
%
%
To train DP LLMs, \textit{Differentially Private Stochastic Gradient Descent (DPSGD)}~\cite{abadi2016deep} is typically used. DPSGD computes gradients for individual data points, clips each individual gradient, and adds Gaussian noise to the average of clipped gradients to ensure $(\epsilon, \delta)$-DP.

\paragraph{Knowledge Distillation: }
In Knowledge Distillation (KD), a small student model is trained to replicate the behavior of a larger teacher model. The student model learns by imitating the outputs of the teacher model. 
Our approach utilizes a variant known as Impossible Distillation~\cite{jung2023impossible} which leverages smaller, pre-trained models to generate high-quality training data.

\newcommand{\wrong}[1]{\textcolor{red}{#1}}
\begin{table*}[t]
\begin{adjustbox}{width=0.95\textwidth,center}
\small
\centering
\renewcommand{\arraystretch}{1.2}
\begin{tabular}{>{\raggedright\arraybackslash}p{1cm} >{\raggedright\arraybackslash}p{2.8cm} >{\raggedright\arraybackslash}p{4cm} >{\raggedright\arraybackslash}p{4.5cm} >{\raggedright\arraybackslash}p{3cm}}
\toprule
\textbf{Category} & \textbf{Type} & \textbf{Definition} & \textbf{Example} & \textbf{Explanation} \\
\midrule
\multirow{4}{*}[-6.5em]{ \rotatebox[origin=c]{90}{\textbf{Inconsistencies}}} &
Extrinsic information &
Summary contains new information not grounded in the source document &
\textit{settling companies..., he added, \wrong{noting that the companies had not yet settled.}} &
Source does not mention whether companies had settled \\
\cmidrule(lr){2-5}
&
Mis-referencing (quote or attribution) &
Property or event found in source, but incorrectly quoted or associated with wrong entity &
\textit{A tropical storm is expected to make landfall in the Gulf of Mexico on Monday night, \wrong{the National Hurricane Center (NHC) has said}.} &
Source does not attribute prediction to the NHC \\
\cmidrule(lr){2-5}
&
Mis-referencing (name or entity) &
Person, property, or event found in source, but associated with wrong name or entity &
\textit{British Cycling has apologised for "failings" in the organisation's World Class programme, according to former world champion \wrong{Laura} Houvenaghel.} &
Houvenaghel's first name is Wendy, not Laura \\
\cmidrule(lr){2-5}
&
Contradiction &
Summary contradicts the source material &
\textit{The Dow Jones Industrial Average .DJI > closed at a record high on Tuesday, \wrong{down 0.2 percent}.} &
Source indicates the market went up, not down \\
\midrule
\multirow{4}{*}[-4.5em]{\rotatebox[origin=c]{90}{\textbf{Language Errors}}} &
Duplicated input &
Model's output is identical to the input &
\textit{\wrong{According to the federal Centers for Disease Control and Prevention (news-web sites)...}} &
Generated sentence is identical to input \\
\cmidrule(lr){2-5}
&
Grammar error &
Grammatically incorrect output &
\textit{man, 26, arrested and charged  after \wrong{search  properties in the city}} &
Missing word ``of'' makes sentence non-grammatical \\
\cmidrule(lr){2-5}
&
Incomplete thought &
Output begins describing a thought or concept but does not complete it &
\textit{Queen's will cut the number of students \wrong{and the staff it.} The cuts will come...} &
Sentence ends without completion \\
\cmidrule(lr){2-5}
&
Missing punctuation &
Output missing clearly required punctuation &
\textit{More than 100 people have been killed in floods in the state \wrong{Gujarat}} &
Sentence ends without period \\
\cmidrule(lr){2-5}
&
Spelling mistake &
Output with at least one spelling mistake &
\textit{East Sussex NHS NHS has \wrong{apo} to patients who were sent leaflets in the mistake} &
Incorrect spelling of ``apologized'' \\
\bottomrule
\end{tabular}
\end{adjustbox}
\caption{Taxonomy and examples of generation errors in language models, categorized into two main types: (1) Inconsistencies (hallucinations) - factual errors including extrinsic information, misattributions, and contradictions, and (2) Language Errors - structural issues such as grammar mistakes, incomplete sentences, and spelling errors. Each error type is illustrated with a representative example (errors highlighted in \textcolor{red}{red}) and detailed explanation. Inconsistency categories are adapted from \citet{tang2024tofueval}.}
\label{tbl:hallucination_examples}
\end{table*}

\section{Proposed Method} 
We present DPRefine, depicted in Figure~\ref{fig:conceptual_understanding_tool}, a three-phase method designed to enhance the linguistic quality of differentially private large language models for domain-specific tasks like summarization and paraphrasing. Our method integrates data synthesis, differentially private fine-tuning, and self-distillation to produce high-quality outputs while maintaining strong privacy guarantees. Our method integrates three key phases: data synthesis, differentially private fine-tuning, and self-distillation. In the first phase, data synthesis provides better initialization and richer feature representations, allowing the model to learn key patterns without privacy concerns. In the second phase, differentially private fine-tuning preserves these learned features while ensuring privacy, adding noise selectively to maintain model robustness. Finally, in the third phase, self-distillation combined with careful filtering further refines the model’s outputs, correcting privacy-induced errors and boosting linguistic quality, all while maintaining strong privacy guarantees.


\subsection{Phase 1: Data Synthesis and Model Initialization}
As depicted in Figure~\ref{fig:conceptual_understanding_tool} (left section), DPRefine begins by generating high-quality input-output pairs \(\{ (a, b), (a, c), \dots , (x, y) \}\) using a small pre-trained language model (e.g GPT-2 for general tasks or BioGPT for biomedical tasks). We choose a smaller model because it offers a balance between efficiency and quality, allowing for fast generation of synthetic data without the computational overhead of larger models. This synthetic data serves as the foundation for fine-tuning the base T5-large model, \(M_{base}\). 

We begin by generating a \textbf{context} \(c\) based on a domain-specific \textbf{prefix}. The prefix ensures alignment with the target domain and can either be generated by a language model or sourced from a human-written corpus to ensure meaningful data. For example, using the prefix \textit{"NYC (Reuters) --"} for the news domain, the model generates a context; \textit{c = "The mayor announced a new climate initiative."}\
Multiple sentence completions \(\{ a, b, c, \dots \}\) are then generated based on the context \(c\). 
(\(a\)) \textit{-"The initiative focuses on creating new parks, reducing emissions, and implementing stricter environmental regulations to combat climate change."} , (\(b\))- \textit{"The plan includes new parks and emission controls."} , \((c\))-\textit{"New parks and stricter emission laws are planned."}...

These generated completions \(\{ a, b, c, \dots \}\) are then paired to form input-output pairs \((x, y)\), where \(x\) is an input and \(y\) is the corresponding output. For instance, potential pairs could be \[(x, y) = (a, b), (a, c), (b, c)\] providing a diverse set of synthetic input-output pairs for training.
 
To ensure high-quality and meaningful pairs, we apply the following filters:

\begin{enumerate}[leftmargin=12pt, itemsep=0pt, topsep=0pt, partopsep=0pt, parsep=0pt]

    \item \textbf{Entailment Filtering:} Using a pre-trained NLI model~\cite{liu2022wanli}, we ensure that the generated pair \((x, y)\) holds logical entailment in both directions, meaning $x \rightarrow y$ and $y \rightarrow x$.
    \item \textbf{Length Filtering:} Ensure the length of the response $y$ is appropriate for the input $x$. For summarization, $y$ should be shorter than $x$ ,  \( |y| < |x| \); for paraphrasing, $x$ and $y$ should have similar length, \(|x| \approx |y|\). 
    
    \item \textbf{Diversity Filtering:} Remove pairs that are too similar to each other to ensure a diverse dataset. Pairs $(x_1,y_1)$  and $(x_2,y_2)$ are duplicates if one pair entails the other: $x_1 \rightarrow x_2$ and $y_1 \rightarrow y_2$.
    
    \item \textbf{Grammar Filtering}: Apply the \texttt{language-tool-python} library~\cite{language_tool_python} to check for grammatical correctness in both \(x\) and \(y\). Any pairs with significant grammatical errors are removed.

    \item \textbf{Numerical Consistency Filtering}: For pairs containing numerical data, ensure that numbers appearing in \(x\) are consistent with those in \(y\), ensuring no significant deviations between the input and output. 

    \item \textbf{Redundancy Filtering:} Remove pairs where more than 30\%  of the tokens in \(y\)  are repeated from \(x\), ensuring minimal redundancy in the generated text.
    \end{enumerate}

These customized filters inspired by the principles in Impossible Distillation~\cite{jung2023impossible}, designed to ensure that the generated samples are highly suited for tasks like summarization and paraphrasing. The filters also counteract common errors exacerbated by DPSGD, such as language errors, as identified in our manual error analysis (see Table~\ref{tbl:hallucination_examples}). By addressing these task-specific and error-related challenges, DPRefine produces cleaner, more accurate outputs for downstream tasks.

After filtering, we compile the curated dataset \(D_{base}\) and fine-tune \(M_{base}\), effectively distilling the knowledge from the pre-trained model to prepare it for specialized tasks like paraphrasing or summarization.

\subsection{Phase 2: DP Task-Specific Fine-Tuning}
In this phase (see middle section in Figure~\ref{fig:conceptual_understanding_tool}), we fine-tune \(M_{base}\) using DPSGD~\cite{abadi2016deep} on a private dataset while ensuring differential privacy. We assume access to this private dataset, which contains sensitive, domain-specific data relevant to the task at hand such as medical text for summarizing biomedical data. This private dataset allows the model to specialize in the target domain while ensuring differential privacy.

This fine-tuning process results in a differentially private model \(M_{private}\), which preserves privacy under \((\epsilon, \delta)\)-DP. The post-processing property of DP ensures that any further use of \(M_{private}\) maintains the same privacy guarantee. The private dataset used in this phase consists of domain-specific data, allowing \(M_{private}\) to specialize while ensuring privacy guarantees.

\subsection{Phase 3: Self-Distillation Refinement}
As shown in Figure~\ref{fig:conceptual_understanding_tool} (right section), the final phase of DPRefine involves using self-distillation to further refine the model. Here, \(M_{private}\) generates new outputs based on input contexts and self-corrects using its own predictions. For each context \(c\), \(M_{private}\) generates multiple output completions \(\{ a', b', c', \dots \}\), which are then paired to form new input-output pairs: \((x', y') = ((a',b'), (b',c'), (a', c'))\). 
The same filtering criteria from Phase 1 are applied to ensure high-quality pairs. After filtering, the data set \(D_{refined}\)  is used to fine-tune \(M_{private}\), resulting in the final model \(M_{refined}\), which balances privacy and output quality.

\begin{algorithm}[ht]
\caption{DPRefine}
\label{alg:DPRefine}
\textbf{Input:} Pre-trained language model \(M_{pre}\), dataset \(D\), privacy parameters \((\epsilon, \delta)\), learning rate \(\eta\), gradient clipping norm \(C\) \\
\textbf{Output:} Final refined model \(M_{refined}\)
\begin{algorithmic}[1]
    \State \textbf{Phase 1: Data Synthesis \& Base Model Training}
    \State Generate a set of contexts \( \{ c_i \}_{i=1}^{N} \), where each \( c_i \) is generated by a language model (LM) using a domain-specific prefix.
    \State For each \( c_i \), generate completions \( \{a, b, c, \dots \} \), where each letter represents a distinct completion for that context.
    \State Form input-output pairs \( \mathcal{P} = \{(a, b), (b, c), \dots \} \) for each context.
    \State Define a filtering function \( \mathcal{F}(\mathcal{P}) \rightarrow \mathcal{P}_{filtered} \), applying filters: entailment, length, diversity, grammar, numerical consistency, and redundancy.
    \State Fine-tune the pre-trained model \( M_{pre} \) on the filtered dataset \( \mathcal{P}_{filtered} \), resulting in \( M_{base} \).

    \Statex

    \State \textbf{Phase 2: DP Fine-Tuning with DPSGD}
    \State Compute noise multiplier \(\sigma\) based on \((\epsilon, \delta)\).
    \For{each minibatch \( B_t \subseteq D \) from the private dataset}
        \State Compute gradients \( g_t = \nabla \mathcal{L}(M_{base}; B_t) \).
        \State Clip the gradient: \( \hat{g}_t = \frac{g_t}{\max(1, \frac{\| g_t \|_2}{C})} \).
        \State Add noise: \( \tilde{g}_t = \hat{g}_t + \mathcal{N}(0, \sigma^2 C^2) \).
        \State Update model: \( M_{private} \leftarrow M_{private} - \eta \tilde{g}_t \).
    \EndFor

    \Statex

    \State \textbf{Phase 3: Self-Distillation Refinement}
    \State Generate new completions \( \{a', b', c', \dots \} \) from \( M_{private} \) for each input context.
    \State Form new input-output pairs \( \mathcal{P}' = \{(a', b'), (b', c'), \dots \} \).
    \State Apply the same filtering \( \mathcal{F}(\mathcal{P}') \) to form \( \mathcal{P}'_{filtered} \).
    \State Fine-tune \( M_{private} \) on \( \mathcal{P}'_{filtered} \), resulting in the final model \( M_{refined} \).

    \Statex
    \State \Return \( M_{refined} \)
\end{algorithmic}
\end{algorithm}

\section{Experimental Setup}
\subsection{Dataset, Models and Data}
\paragraph{Datasets and models.} We evaluated DPRefine across three datasets for two domain-specific tasks: summarization and paraphrasing. For summarization, we used the XSum dataset~\cite{narayan2018don}, containing 204,045 training samples and 11,334 test samples. XSum consists of BBC articles paired with single-sentence summaries, making it particularly challenging due to the requirement for highly concise yet informative summaries. We also used the PubMed dataset~\cite{pubmed}, which contains 119,924 training samples and 6,658 test samples of biomedical research articles with structured abstracts. This dataset demands the summarization of technical and domain-specific content. For paraphrasing, we utilized the MRPC dataset from the GLUE benchmark~\cite{wang2018glue}, which contains 3,668 training samples and 1,725 test samples of sentence pairs automatically extracted from news articles, with human annotations for semantic equivalence. The smaller size of MRPC makes it particularly well-suited for tasks requiring careful paraphrasing and semantic retention.

To generate synthetic data, we used GPT-2~\cite{radford2019language} for general tasks like XSum and MRPC, while BioGPT~\cite{luo2022biogpt} was employed for domain-specific biomedical text in PubMed. These smaller models were chosen due to their efficiency in producing high-quality data quickly, without the computational overhead of larger models. The synthetic data generated by these models was then used to fine-tune the base paraphrasing/summarization model, T5-large, referred to as \(M_{base}\). While the training data for GPT-2 and T5-large is not publicly available, we use XSum, PubMed, and MRPC as proxies for sensitive data, ensuring transparency and reproducibility.

\paragraph{Baselines.} We compared DPRefine against several baselines. The non-private baselines included Copy-Input, GPT-4, and T5-large. Copy-Input provides a simplistic baseline for paraphrasing tasks by replicating the input directly. GPT-4 represents an upper bound for performance without privacy constraints, and T5-large was fine-tuned directly on the datasets without privacy mechanisms, serving as a middle-ground non-private baseline. For private baselines, we applied DPSGD to T5-large (DPSGD\textsubscript{T5}) to evaluate how much our approach improves over traditional differentially private training.


\subsection{Implementation Details} 

\textit{\paragraph{Synthetic Data Generation:}} In Phase 1, we used GPT-2 and BioGPT to generate synthetic data with nucleus sampling (top-p=0.9, temperature=0.1) and a token limit of 150. Contextual prefixes (e.g., "New York (Reuters) –") were used for XSum and MRPC to enhance diversity, while BioGPT generated PubMed data without prefixes.  

\textit{\paragraph{Filtering:}} We applied several filters to improve the generated data. First, length filtering ensured the output text was concise ($\leq75\%$ of the input length). We then applied semantic equivalence filtering using RoBERTa-large-WANLI, retaining only pairs with a bidirectional entailment score above 0.95. Reverse-NLI filtering ensured logical consistency between input and output, with a threshold of 0.7. Additional filters removed redundant tokens ($\leq30\%$ repetition), ensured numerical consistency, and checked for grammatical errors using the language-tool-python library. Finally, a graph-based approach identified and removed duplicate paraphrases based on entailment scores. 

\textit{\paragraph{Model Training:}} Both the base model, \(M_{base}\) and the final model \(M_{refined}\) were trained using the same configuration. We used T5-large with an AdamW optimizer, a learning rate of $5e^{-5}$, gradient clipping set at 1.0, and Perplexity (PPL) as the main evaluation metric. Training was conducted over 1 epoch with a batch size of 8 for training and 16 for validation. Beam search and top-p sampling (top-p = 0.9) ensured output diversity.

\textit{\paragraph{DP Fine-Tuning:}} In Phase 2, using the private-transformers codebase~\cite{privatetransformers} we fine-tuned the T5-large model (\(M_{base}\) using DPSGD. For each dataset, we set a privacy budget $\epsilon$ to 8,  $\delta = \frac{1}{2N}$ and a clipping parameter of 1.0. After basic hyperparameter tuning, training was conducted over four epochs with a batch size of 4 with a learning rate of $5e^{-5}$, and batch size of 4 resulting in \(M_{private}\).

\textit{\paragraph{Self Distillation:}} In Phase 3, \(M_{private}\) generated new outputs using inputs from Phase 1 data. These outputs were filtered and used to further fine-tune \(M_{private}\), resulting in \(M_{refined}\), following the same training setup as in earlier phases.



\begin{figure*}
    \centering
    \begin{subfigure}[b]{0.31\textwidth} 
        \centering
        \includegraphics[width=\textwidth]{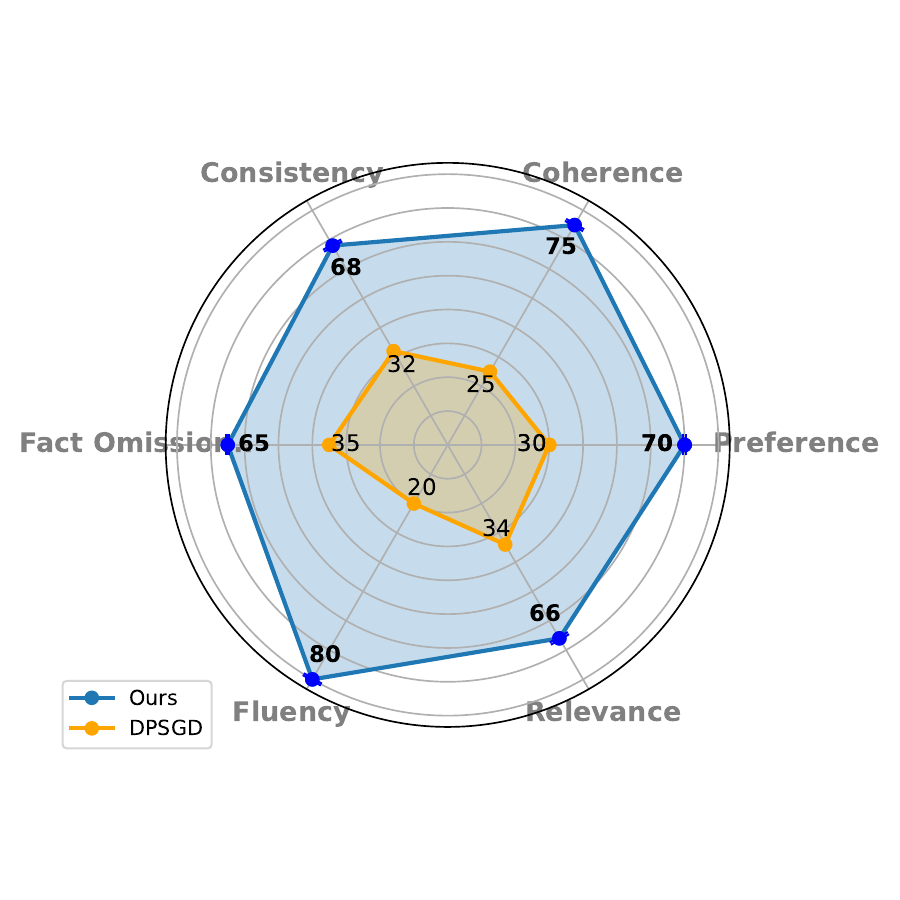}
        \vspace*{-40pt}
        \caption{XSum Dataset}
        \label{subfig:xsum}
    \end{subfigure}%
    \hfill 
    \begin{subfigure}[b]{0.31\textwidth}
        \centering 
        \includegraphics[width=\textwidth]{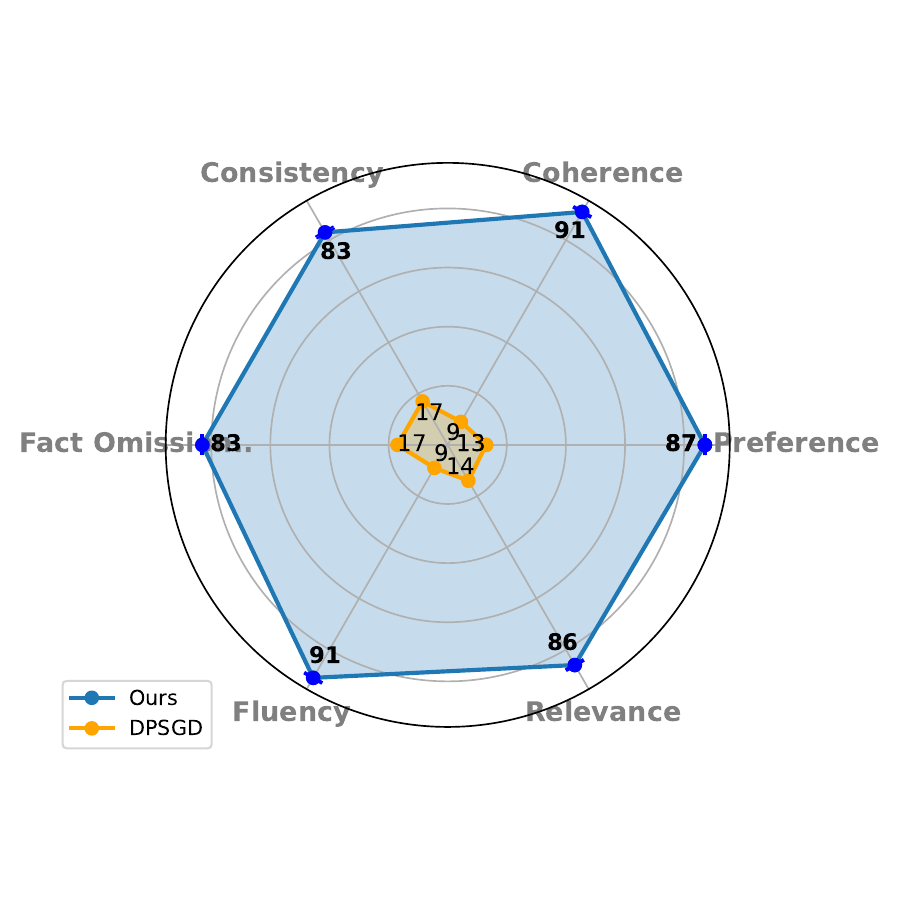}
        \vspace*{-40pt}
        \caption{PubMed Dataset}
        \label{subfig:pubmed}
    \end{subfigure}
    \hfill
    \begin{subfigure}[b]{0.31\textwidth}
        \centering 
        \includegraphics[width=\textwidth]{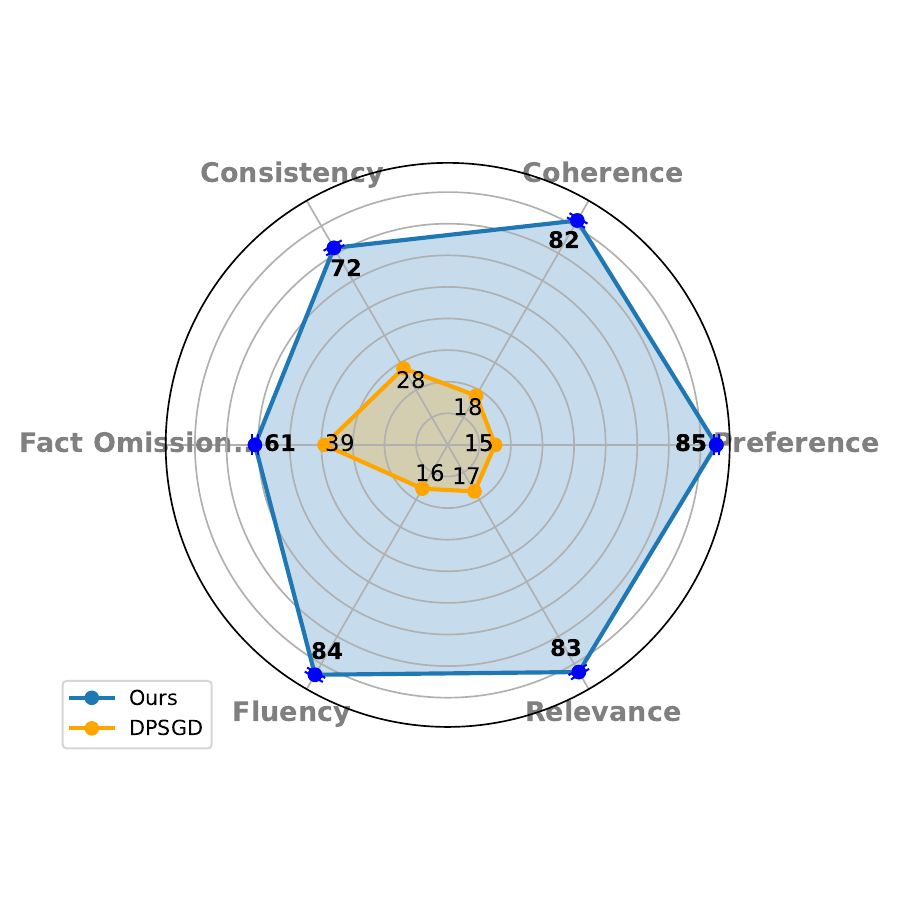} 
        \vspace*{-40pt}
        \caption{MRPC Dataset}
        \label{subfig:mrpc}
    \end{subfigure}
    \caption{ Comparison of DPRefine and DPSGD across multiple metrics (Preference, Coherence, Consistency, Fact Omission, Fluency, and Relevance) for the XSum, PubMed, and MRPC datasets. Error bars represent standard deviations. DPRefine demonstrates consistently stronger performance showing its ability to generate more contextually aligned and factually accurate outputs compared to DPSGD.}
    
    \label{fig:alpaca_results}
    \vspace{-2ex}
\end{figure*}

\section{Experimental Results} 
In this section we evaluate the efficacy of DPRefine. We conduct extensive quantitative and qualitative analysis: (1) overall evaluation using LLM-as-a-judge, (2) reference-based, targeted assessments and  (4) diversity evaluations.  Then, we provide a fine-grained, qualitative error-analysis which categorizes the types of errors different models make, and exemplifies the improvements that DPRefine provides. Finally, we perform ablation studies. 

\subsection{LLM-as-a-judge Evaluation}
\paragraph{Setup:} We evaluated DPRefine and DPSGD using AlpacaEval~\cite{li2023alpacaeval}, performing pairwise comparisons on the test sets of all datasets. \textit{GPT-4-1106-preview} was used to assess the models across six metrics: preference, coherence, consistency, fact omission, fluency, and relevance. The evaluation was based on modified prompts(see Appendix \ref{sec:appendix_alpcal}) tailored to measure the quality of generated outputs on these dimensions.  To ensure a fair comparison, we include both non-private and private baselines. The non-private baselines (e.g., GPT-4, T5-large) demonstrate the upper bounds of performance without privacy constraints, while private baselines (e.g., DPSGD) allow us to compare how our approach improves over traditional differentially private models.

\paragraph{Results:} As shown in Figure \ref{fig:alpaca_results}, DPRefine consistently outperforms DPSGD across all evaluation metrics. On average, AlpacaEval preferred DPRefine’s generations in 78.38\% of cases across all datasets and metrics. The model shows higher scores in relevance and consistency, indicating better alignment with the input and fewer contradictions. While fluency was slightly lower, DPRefine's superior performance in coherence and fact omission suggests more logically structured and accurate outputs. These results demonstrate that DPRefine generates outputs with stronger contextual and factual alignment compared to DPSGD. These findings are further supported by our manual analysis, which shows that DPRefine significantly reduces inconsistencies and language errors compared to DPSGD, validating the LLM-as-judge results.

\subsection{Targeted Reference-based Evaluation} 
\paragraph{Setup:} For reference-based metrics, we evaluate DPRefine using ROUGE-L \cite{lin2004rouge} and BERT-F1 \cite{zhang2019bertscore} to assess token overlap and semantic similarity with reference outputs. Additionally, we use iBLEU \cite{sun2012joint} and BERT-iBLEU (B-iB), which offer a more comprehensive assessment of output quality, with BERT-iBLEU being particularly useful for capturing semantic preservation, as it correlates better with human judgments than token-based metrics \cite{niu2020unsupervised}. iBLEU is a variant of the BLEU metric that balances adequacy and diversity by rewarding similarity to the reference output while penalizing excessive overlap with the input, making it well-suited for paraphrasing tasks. Given that our tasks include summarization and paraphrasing, we use these automated metrics to measure performance against ground truth outputs. We compare both non-private and private baselines to evaluate DPRefine’s performance under privacy constraints.


\paragraph{Results:} 
Table \ref{table:reference_based_metrics} shows that Copy-Input, a simple baseline that copies the input directly, achieves high scores on ROUGE-L and BERT-F1, indicating that token overlap metrics often favor models that produce outputs closely resembling the input especially for paraphrasing tasks. However, BERT-iBLEU highlights the limitations of Copy-Input, as it performs poorly on semantic similarity with the reference paraphrase. DPRefine outperforms DPSGD across iBLEU and BERT-iBLEU, particularly in MRPC (80.55 in BERT-iBLEU) and PubMed (75.55 in BERT-iBLEU), reflecting its ability to generate semantically accurate outputs. 
While privacy-preserving models naturally experience some performance trade-offs compared to non-private models, DPRefine maintains high semantic accuracy and fluency even in privacy-constrained settings. Although DPRefine’s scores on ROUGE-L and BERT-F1 are slightly lower than DPSGD in some tasks, its superior performance on BERT-iBLEU suggests that it captures deeper semantic meaning.

These results indicate that reference-based metrics like ROUGE may not fully reflect the improvements in output quality brought by DPRefine, particularly in tasks requiring greater semantic accuracy. This aligns with prior findings ~\cite{cohan2016revisiting, liu2008correlation, reiter2009investigation, stent2005evaluating} that traditional metrics often fail to evaluate the true quality of summaries, especially in datasets like XSum where reference summaries are of questionable quality \cite{zhang2024benchmarking}. 

\subsection{Diversity Evaluation} 

\paragraph{Setup:} For lexical diversity, we follow Liu et al. to compute Lexical Deviation (LD) and Word Pair Deviation (WPD), which assess the degree of variation between input and output~\cite{liu2022towards}. We also measure vocabulary richness using the mean-segmented token type ratio (MSTTR)~\cite{torruella2013lexical} and the token-level Jaccard similarity between source and
predicted output.


\paragraph{Results:} DPRefine performs consistently well in MSTTR and Jaccard Similarity, slightly exceeding DPSGD across most datasets, with only a marginal difference in PubMed. This indicates that DPRefine improves vocabulary richness and reduces word overlap more effectively than DPSGD, though the improvements are modest. These results suggest that DPRefine introduces incremental enhancements in lexical diversity compared to DPSGD, especially in general tasks.

DPRefine's most significant gains are observed in WPD and LD, particularly in the MRPC paraphrasing task. In MRPC, DPRefine achieves significantly higher scores in both LD and WPD, demonstrating its ability to introduce greater structural variation and diversity between input and output. This is especially important for paraphrasing tasks, where generating diverse sentence structures is essential for varied and meaningful outputs.

\begin{figure*}
    \centering
    \begin{subfigure}[b]{0.34\textwidth} 
        \centering
        \includegraphics[width=\textwidth, trim={0 0 0 10pt}, clip]{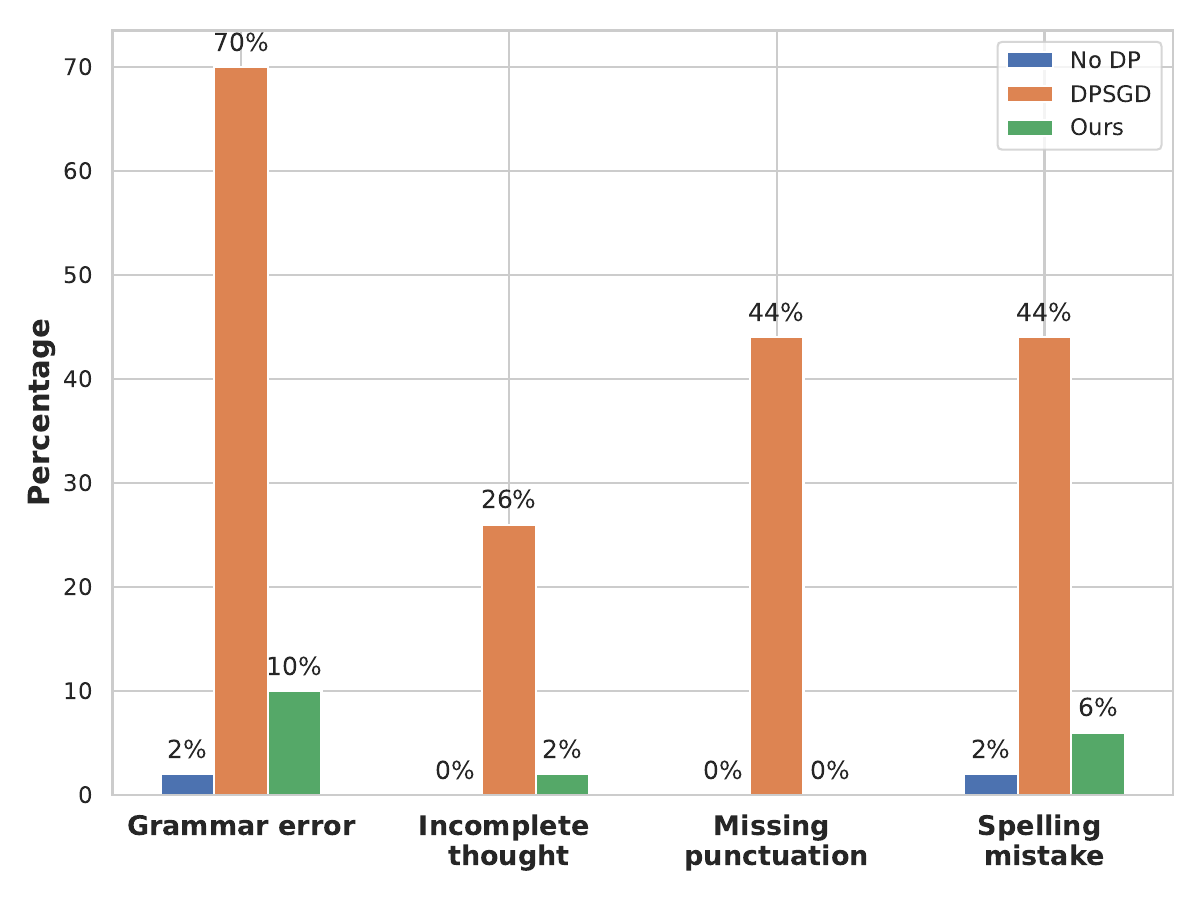}
        \caption{XSum - Language Errors}
        \label{subfig:xsum_language}
    \end{subfigure}%
    \hspace{-0.8em} 
    \begin{subfigure}[b]{0.34\textwidth}
        \centering 
        \includegraphics[width=\textwidth, trim={0 0 0 10pt}, clip]{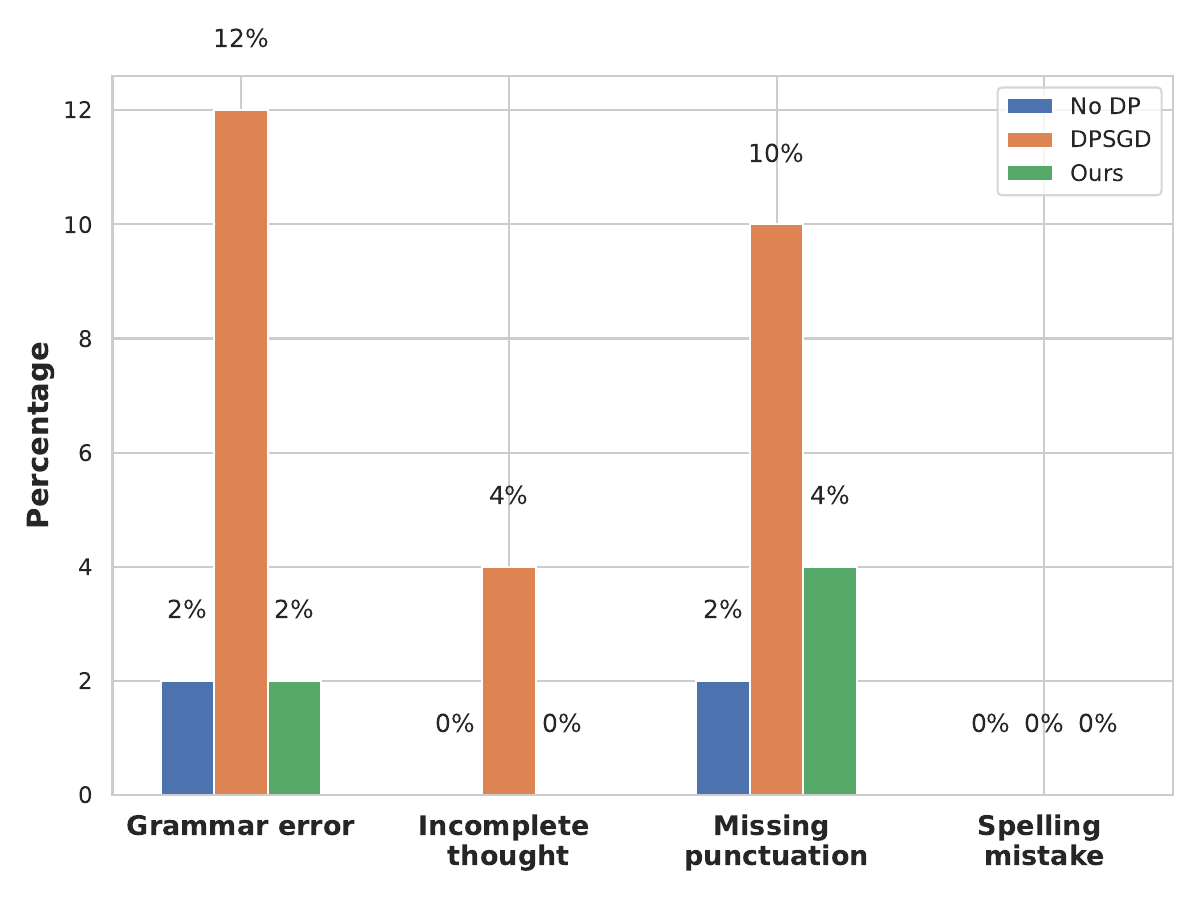}
        \caption{MRPC - Language Errors}
        \label{subfig:mrpc_language}
    \end{subfigure}%
    \hspace{-0.8em} 
    \begin{subfigure}[b]{0.33\textwidth}
        \centering 
        \includegraphics[width=\textwidth, trim={0 0 0 10pt}, clip]{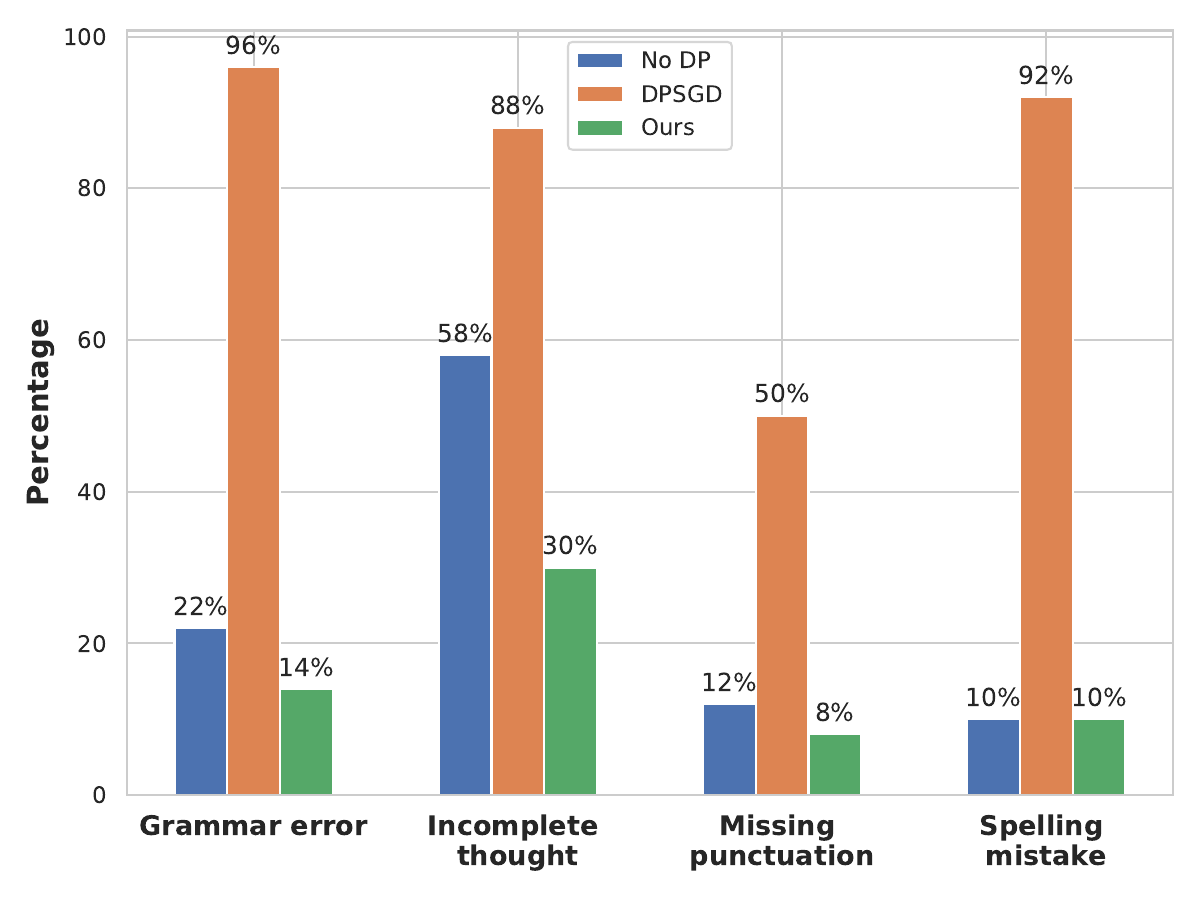}
        \caption{PubMed - Language Errors}
        \label{subfig:pubmed_language}
    \end{subfigure}
    
    \vspace{0.8em} 
    \begin{subfigure}[b]{0.34\textwidth}
        \centering 
        \includegraphics[width=\textwidth, trim={0 0 0 5pt}, clip]{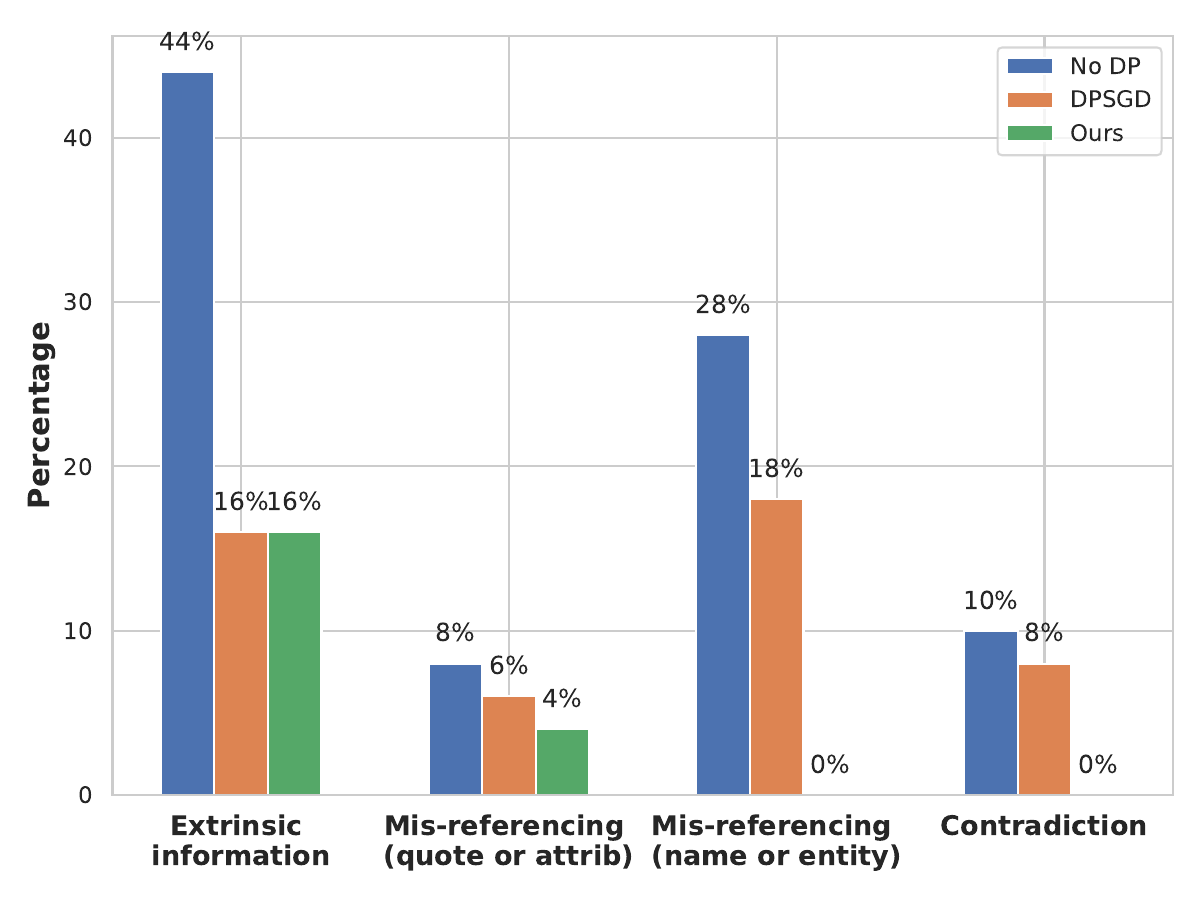}
        \caption{XSum - Inconsistency Errors}
        \label{subfig:xsum_inconsistency}
    \end{subfigure}%
    \hspace{-0.8em}
    \begin{subfigure}[b]{0.34\textwidth}
        \centering 
        \includegraphics[width=\textwidth, trim={0 0 0 5pt}, clip]{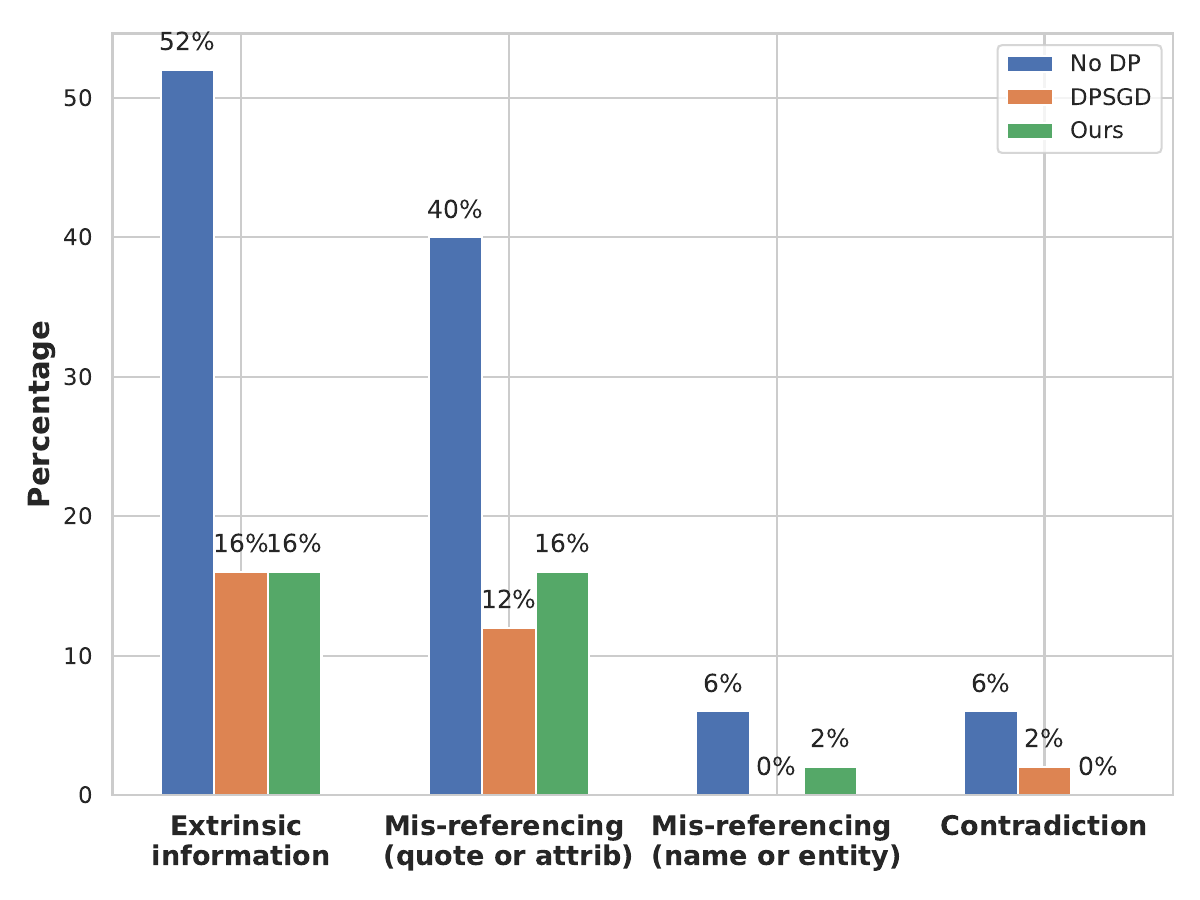}
        \caption{MRPC - Inconsistency Errors}
        \label{subfig:mrpc_inconsistency}
    \end{subfigure}
    \hspace{-0.8em} 
    \begin{subfigure}[b]{0.33\textwidth}
        \centering 
        \includegraphics[width=\textwidth, trim={0 0 0 10pt}, clip]{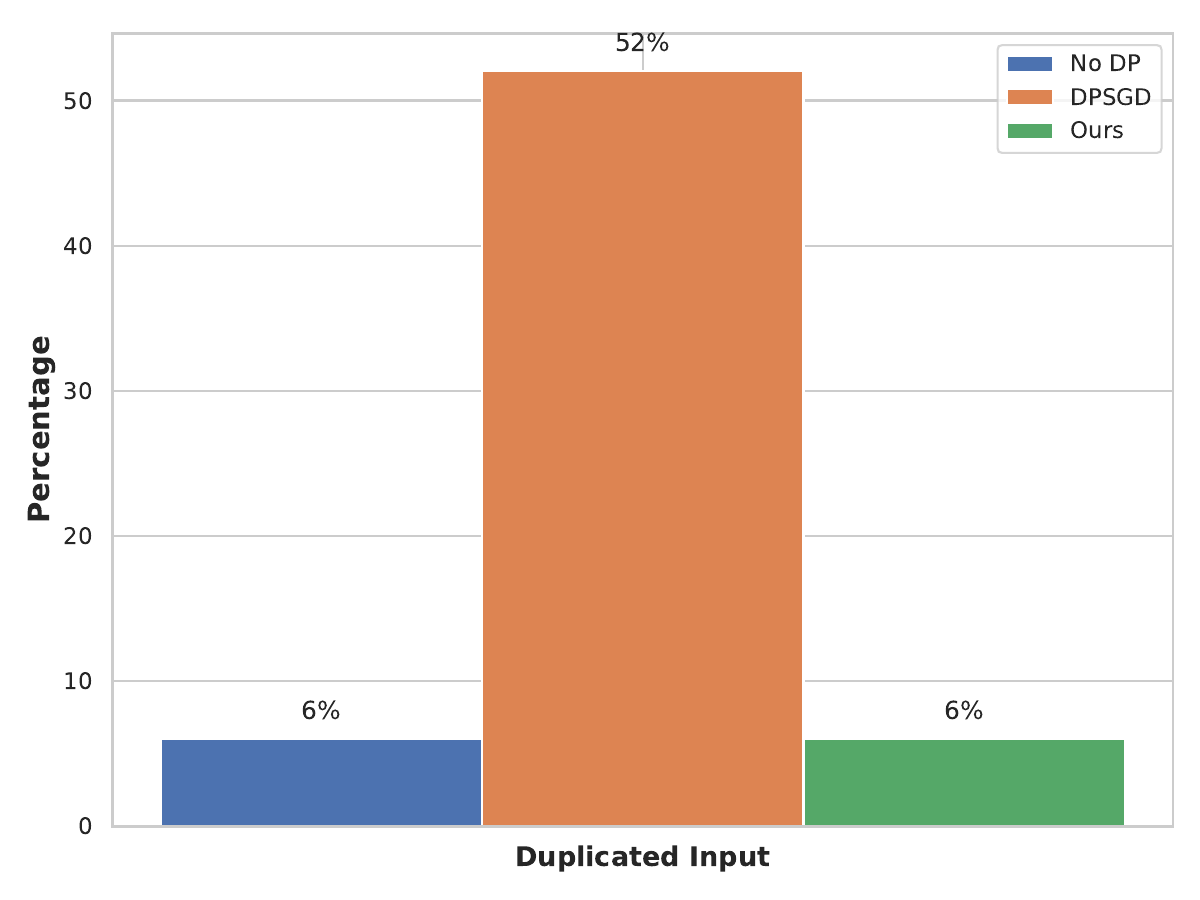}
        \caption{MRPC - Duplicated Input Errors}
        \label{subfig:pubmed_language2}
    \end{subfigure}
    
    \caption{Manual analysis of error types for three models: No DP, DPSGD, and DPRefine across XSum, MRPC, and PubMed datasets. The first row shows language errors, and the second row shows inconsistency errors. DPRefine consistently reduces both language errors and inconsistencies compared to DPSGD, leading to more accurate and fluent outputs. }
    \label{fig:error_analysis_grid}
\end{figure*}

\subsection{Qualitative Error Analysis}

To investigate the impact of DPSGD and distillation on inconsistencies and linguistic quality, we manually annotated 50 outputs each from the XSum, MRPC, and Pubmed test sets. We annotated only language errors in the Pubmed results, since the authors lack the necessary domain knowledge to evaluate inconsistencies for the associated task.
%
To categorize inconsistencies and language errors, we developed the taxonomy shown in Table~\ref{tbl:hallucination_examples}. For inconsistencies and hallucinations, we adopt the types and definitions proposed by Tang et al.~\cite{tang2024tofueval}. For language errors, we developed our taxonomy based on the errors observed in the model outputs. The results of our qualitative analysis appear in Figure~\ref{fig:error_analysis_grid}.

\paragraph{Inconsistencies.}
Our results (Figures~\ref{subfig:xsum_inconsistency} and~\ref{subfig:mrpc_inconsistency}) show that non-private models introduce all kinds of inconsistencies listed in Table~\ref{tbl:hallucination_examples}, and {both DPSGD and \ours \emph{reduce these inconsistencies} significantly}.

In both the summarization and paraphrasing tasks, the non-private model introduces \emph{extrinsic information} in nearly 50\% of all outputs; both DP methods reduce these inconsistencies to 16\% for both tasks. In the paraphrasing task (MRPC), the non-private model \emph{mis-references quotes} about 40\% of the time, and \emph{mis-references entities} about 6\% of the time; both DP methods reduce these inconsistencies significantly. The non-private model introduces \emph{contradictions} in 10\% (XSum) and 6\% (MRPC) of cases; both DP methods reduce them.

\paragraph{Language errors.}
Our results (Figures~\ref{subfig:xsum_language}, \ref{subfig:mrpc_language}, and \ref{subfig:pubmed_language}) show that DPSGD introduces language errors of all types at significant rates, and that \ours \emph{reduces language errors to nearly the level of the non-private model}. On the PubMed dataset---the most challenging task in our evaluation---\ours consistently introduced fewer language errors than even the non-private model. In the paraphrasing task (MRPC), DPSGD often duplicated the input; \ours eliminated this behavior. See Appendix~\ref{sec:appendix_dpsgd} for more discussion, and Appendix~\ref{sec:appendix_hallucination} for example outputs from all models.

\subsection{Ablation Studies}
\subsubsection{Disentangling The Impact of Each Phase} We systematically evaluated the impact of each phase in DPRefine by comparing the performance of three models: \(M_{base}\), \(M_{private}\), and \(M_{refined}\). These correspond to models trained on synthetic data (Phase 1), fine-tuned with differential privacy (Phase 2), and refined via self-distillation (Phase 3), respectively. 

\paragraph{Results:} As shown in Table \ref{tbl:ablation_phases}, \(M_{base}\) (trained on synthetic data) demonstrates strong diversity but lacks consistency and fluency due to the limitations of training on synthetic data alone. Fine-tuning with DPSGD (\(M_{private}\)) improves fluency and semantic accuracy, though at the cost of some diversity due to the privacy noise. Self-distillation (\(M_{refined}\)) recovers much of the lost diversity while maintaining or improving semantic accuracy across datasets, particularly in MRPC and XSum.The results confirm that each phase plays a critical role in balancing privacy and output quality. Removing any phase leads to a reduction in performance, underscoring the necessity of the full DPRefine pipeline. See Appendix \ref{phase_output} for qualitative examples.

\subsubsection{Privacy-utility Trade-off} 
To explore the impact of differential privacy on output quality, we conducted experiments using two privacy budgets: $\epsilon = 3$ and $\epsilon = 8$. For each $\epsilon$, we measured the sampling efficiency—defined as the ratio of outputs that passed our quality filters—as well as the semantic accuracy.

The results in Table \ref{tbl:epsilon_ablation} show that, at $\epsilon = 3$, the model's performance declines in both sampling efficiency and output quality compared to $\epsilon = 8$. For instance, in the MRPC dataset, DPRefine shows lower BERT-iBLEU scores at $\epsilon = 3$. However, with $\epsilon = 8$, both DPRefine and DPSGD produce higher-quality outputs, with DPRefine outperforming DPSGD.

The sampling efficiency results further emphasize the trade-off: a lower $\epsilon$ value introduces more noise, reducing the number of high-quality outputs. At $\epsilon = 8$, DPRefine not only improves utility but also generates a higher proportion of usable outputs (e.g., 51.7\% in MRPC compared to 46.93\% at $\epsilon = 3$), demonstrating that less noise allows the model to retain more valuable content, balancing privacy and output quality.

\begin{table*}[ht]
\centering
\begin{adjustbox}{width=\textwidth,center}
\small
\begin{tabular}{llcccccccccccc}
\toprule
& & \multicolumn{4}{c}{\textbf{XSUM}} & \multicolumn{4}{c}{\textbf{PUBMED}} & \multicolumn{4}{c}{\textbf{MRPC}} \\
\cmidrule(lr){3-6} \cmidrule(lr){7-10} \cmidrule(lr){11-14} 
& & \textit{iBLEU $\uparrow$} & \textit{B-iB $\uparrow$} & \textit{R-L $\uparrow$} & \textit{BERT-F1 $\uparrow$} & \textit{iBLEU $\uparrow$} & \textit{B-iB $\uparrow$} & \textit{R-L $\uparrow$} & \textit{BERT-F1 $\uparrow$} & \textit{iBLEU $\uparrow$} & \textit{B-iB $\uparrow$} & \textit{R-L $\uparrow$} & \textit{BERT-F1 $\uparrow$} \\
\midrule
\multirow{3}{*}{\begin{tabular}[c]{@{}c@{}}\textbf{Non-}\\\textbf{Private}\end{tabular}}  
& Copy-Input & -  & -  & - & - & - & -  & - & - & 24.36  & 78.75  & \underline{62.08} & \underline{93.80} \\
& GPT-4 & \underline{28.93}  & 73.69 & 13.91 & 86.36 & \underline{32.07}   & \underline{74.97} & \underline{19.44} & \underline{83.54} & 30.91  & \underline{82.57} & 42.61 & 91.35 \\
& T5-large & 35.0  & \underline{79.96} & \underline{31.60} & \underline{90.59} & 31.83  & 73.67 & 18.10 & 83.18 & 30.25  & 78.92 & 54.89 & 91.85 \\
\cmidrule(lr){2-14}
\multirow{2}{*}{\begin{tabular}[c]{@{}c@{}}\textbf{Private}\end{tabular}} 
& DPSGD\textsubscript{T5} & \textbf{30.49}  & 74.69 & \textbf{19.35} & \textbf{87.02} & 30.80   & 71.89 & 15.13 & 81.37 & 28.72  & 78.30 & \textbf{52.43} & \textbf{90.76} \\
& DPRefine\textsubscript{T5}  & 30.10  & \textbf{75.55} & 17.73 & 86.51 & \textbf{31.06}   & \textbf{72.93} & \textbf{16.08} & \textbf{82.10} & \textbf{30.65}  & \textbf{80.55} & 42.49 & 90.56 \\
\bottomrule
\end{tabular}
\end{adjustbox}
\caption{Reference-based evaluation results across XSum, PubMed, and MRPC. The first group lists non-private baselines, including Copy-Input, GPT-4, and T5. The second group presents private models (DPSGD and DPRefine). We highlight the top-performing scores for each model in bold, and underline the best non-private model results.}
\label{table:reference_based_metrics}
\end{table*}


\begin{table}[ht]
\centering
\small
\setlength{\tabcolsep}{3pt}
\begin{adjustbox}{width=\linewidth}
\begin{tabular}{llcccc}
\toprule
& & \multicolumn{2}{c}{\textbf{Lexical Diversity }} & \multicolumn{2}{c}{\textbf{Lexical Deviation}} \\  
\cmidrule(lr){3-4} \cmidrule(lr){5-6}
& \textbf{Model} & \textit{MSTTR} $\uparrow$ & \textit{Jaccard Similarity} $\uparrow$ & \textit{LD} $\uparrow$ & \textit{WPD} $\uparrow$ \\
\midrule
\multirow{3}{*}{\textbf{XSum}}  
& Non-Private & \underline{92.37} & \underline{93.73} & \underline{89.39} & \underline{32.20} \\
& DPSGD & 82.03 & 83.90 & \textbf{87.08} & 29.93 \\
& Ours & \textbf{85.24} & \textbf{87.18} & 84.83 & \textbf{30.47} \\
\cmidrule(lr){2-6}
\multirow{2}{*}{\textbf{PubMed}} 
& Non-Private & \underline{78.74} & \underline{82.49} & \underline{91.73} & \underline{33.70} \\
& DPSGD\textsubscript{T5} & 81.30 & 81.18 & \textbf{92.63} & 31.76 \\
& Ours\textsubscript{T5} & \textbf{81.31} & \textbf{81.36} & 91.49 & \textbf{33.11} \\
\cmidrule(lr){2-6}
\multirow{2}{*}{\textbf{MRPC}} 
& Non-Private & \underline{91.92} & \underline{92.53} & 22.50 & 11.33 \\
& DPSGD\textsubscript{T5} & \textbf{89.93} & 91.26 & 23.63 & 13.47 \\
& Ours\textsubscript{T5} & 89.39 & \textbf{91.91} & \textbf{45.59} & \textbf{18.08} \\
\bottomrule
\end{tabular}
\end{adjustbox}
\caption{Results across multiple datasets for different models, comparing Lexical Diversity (MSTTR and Jaccard Similarity) and Lexical Deviation (LD and WPD).}
\end{table}

\begin{table*}[ht]
\centering
\begin{adjustbox}{width=0.9\textwidth,center}
\small
\setlength{\tabcolsep}{3pt}
\begin{tabular}{llcccc}
\toprule
& & \textbf{iBLEU $\uparrow$} & \textbf{BERT-iBLEU $\uparrow$} & \textbf{MSTTR $\uparrow$} & \textbf{Jaccard Similarity $\uparrow$} \\
\midrule
\multirow{4}{*}{\textbf{XSum}}  
& \(M_{base}\) (Pretrained on Synthetic Data) & 29.92 & 74.0 & \textbf{87.48} & \textbf{89.68} \\
& \(M_{private}\) (DP Fine-tuned) & \textbf{30.42} & 75.04 & 81.79 & 83.96 \\
& \(M_{refined}\) (Self-Distilled) & 30.13 & \textbf{75.55} & 85.20 & 87.18 \\
\cmidrule(lr){2-6}
\multirow{4}{*}{\textbf{PubMed}} 
& \(M_{base}\) (Pretrained on Synthetic Data) & \textbf{32.06} & \textbf{74.97} & 71.87 & 76.18 \\
& \(M_{private}\) (DP Fine-tuned) & 30.86 & 72.33 & 79.09 & 79.36 \\
& \(M_{refined}\) (Self-Distilled) & 31.05 & 72.93 & \textbf{81.05} & \textbf{81.36} \\
\cmidrule(lr){2-6}
\multirow{4}{*}{\textbf{MRPC}} 
& \(M_{base}\) (Pretrained on Synthetic Data) & 30.04 & 80.36 & 87.56 & 90.24 \\
& \(M_{private}\) (DP Fine-tuned) & 30.55 & 80.53 & 86.87 & 89.79 \\
& \(M_{refined}\) (Self-Distilled) & \textbf{30.65} & \textbf{80.55} & \textbf{89.38} & \textbf{91.90} \\
\bottomrule
\end{tabular}
\end{adjustbox}

\caption{Performance comparison across multiple datasets for \(M_{base}\) (Pretrained on Synthetic Data), \(M_{private}\) (DP Fine-tuned), and \(M_{refined}\) (Self-Distilled) models.} 
\label{tbl:ablation_phases}
\end{table*}

\begin{table}
\centering
\small
\begin{adjustbox}{width=\linewidth}
\begin{tabular}{llccc}
\toprule 
& & \multicolumn{2}{c}{\textbf{BERT-iBLEU}} & \textbf{Sampling Efficiency } \\
\cmidrule(lr){3-4} \cmidrule(lr){5-5}
\textbf{Dataset} & \textbf{$\epsilon$} & \textbf{DPSGD} & \textbf{DPRefine} & \textbf{DPRefine} \\
\midrule
\multirow{2}{*}{\textbf{XSum}}  
    & 3 & 74.78 & 75.01 & 28.23\% (847/3000)\\
    & 8 & 74.68 & \textbf{75.55} & 30.4\% (912/3000)\\
\cmidrule(lr){2-5}
\multirow{2}{*}{\textbf{MRPC}}  
    & 3 & 78.25 & 80.54 & 46.93\% (1408/3000) \\
    & 8 & 78.30 & \textbf{80.55} & 51.7\% (1551/3000) \\
\bottomrule
\end{tabular}
\end{adjustbox}
\caption{Comparison of BERT-iBLEU and sampling efficiency for DPSGD and DPRefine models across different datasets at privacy budgets of $\epsilon = 3$ and $\epsilon = 8$.}
\label{tbl:epsilon_ablation}
\end{table}

\section{Related Work}

Our work builds on existing efforts to balance privacy and utility in NLP, particularly through the use of DPSGD for fine-tuning models on both private and public data~\cite{kerrigan2020differentially, li2021large, yu2021differentially, mireshghallah2022privacy, ganesh2023public}. DPSGD has been widely adopted for maintaining privacy guarantees during training, but it often comes at the cost of reduced utility ~\cite{yu2021differentially,ponomareva2022training,bagdasaryan2019differential,mireshghallah2022privacy}.

Knowledge Distillation (KD) has also been used to improve model performance and reduce the size of LLMs\cite{jiao2019tinybert, sun2019patient}. However, combining KD with DP remains under explored, and only a few studies address the utility losses that arise from using both techniques together\cite{mireshghallah2022differentially, xie2024differentially}. Our work seeks to fill this gap by enhancing the linguistic quality of DP-trained LLMs through a combination of KD and self-distillation.

Recent work has also focused on the use of synthetic data to enhance private model training. Flemings et al. introduced DistilDP, which leverages DP synthetic data for knowledge distillation to minimize utility loss in compressed models under privacy constraints~\cite{flemings2024differentially}, while Yu et al. explored DP synthetic data generation for training lightweight models~\cite{yu2023training}. Xie et al. proposed AUG-PE, generating DP synthetic text via API access to large models without fine-tuning~\cite{xie2024differentially}. Unlike these works, DPRefine emphasizes not only model compression and privacy but emphasizes on enhancing linguistic quality and model scalability without relying on large foundational models or relying on API access.

Hallucinations in LLMs are another critical concern, where models generate incorrect or unsubstantiated information. Some recent methods for addressing hallucinations include entropy-based detection~\cite{farquhar2024detecting}, finetuning unfamiliar examples~\cite{kang2024unfamiliar}, and self-reflection techniques to reduce errors in domain-specific tasks~\cite{ji2023towards}. Kang et al. further observed that neural networks tend to extrapolate predictably toward a constant value when faced with out-of-distribution (OOD) inputs, contributing to predictable hallucinations in certain contexts~\cite{kang2023deep}.  DPRefine incorporates these insights to minimize hallucinations while maintaining privacy and output quality.


\section{Conclusion}
We introduced DPRefine, a three-phase method to improve the utility and linguistic quality of differentially private language models. DPRefine addresses DPSGD’s limitations by combining data synthesis, privacy-preserving fine-tuning, and self-distillation, resulting in more accurate, coherent, and diverse outputs. Our experiments show DPRefine outperforms DPSGD, reducing language errors and inconsistencies while preserving privacy. By using synthetic data for initialization and self-distillation for refinement, DPRefine mitigates privacy noise and balances quality with privacy guarantees. Fine-grained analysis confirms that DPRefine significantly reduces inconsistencies and errors compared to DPSGD, making it a promising approach for developing high-performing, privacy-preserving models for sensitive tasks. Future work could explore its application to additional NLP tasks and larger models.


\section{Social Impacts Statement}
In this work, we utilized pre-trained LLMs and well-known language modeling datasets accessed from the Hugging Face API, which are publicly available and free to use. Specifically, we employed GPT-2 and T5 models licensed under the Apache License, Version 2.0. The datasets used in our experiments, including XSum, PubMed, and MRPC, are widely recognized in the academic community and are used under their respective licenses.

Our intended use of these artifacts is strictly for academic research, aligning with the intended use specified by the creators. The datasets were chosen for their relevance to the tasks of summarization and paraphrasing, ensuring that they do not contain personally identifiable information (PII). The PubMed dataset, while containing potentially sensitive information, was handled with care to focus solely on text content without disclosing any personal or identifiable data. 

By utilizing publicly available models and datasets and implementing differential privacy techniques, we minimize any unintended privacy leakage that could result from experimenting. 

\section{Limitations}
Our approach, while showing promising improvements in the linguistic quality of DP LLMs, comes with certain limitations. First, the computational cost of the multi-phase distillation process is significant, particularly in Phase 1 where large amounts of synthetic data are generated and filtered. This might be impractical for settings with limited computational resources. Additionally, the effectiveness of our approach heavily relies on the quality and diversity of the initial synthetic dataset. If the generated data is not representative or diverse enough, the benefits of the subsequent distillation phases may be diminished.

Furthermore, our method has been evaluated on specific datasets and tasks, such as summarization and paraphrasing. The generalizability of our findings to other NLP tasks or different types of datasets remains to be explored. Lastly, while our approach reduces inconsistencies, it does not entirely eliminate them, and some errors still persist, which could affect the reliability of the generated outputs.

\section*{Acknowledgments}

We would like to thank Sean Welleck for helpful discussion at the conception of the idea, and Jaehun Jung for their help with the experimental setup and evaluation choices. We would also like to thank the members of the UVM PLAID lab and UW NLP for helpful discussion and feedback.

This material is based upon work supported by the National Science Foundation under Grant No. 2238442. Any opinions, findings and conclusions or recommendations expressed in this material are those of the author(s) and do not necessarily reflect the views of the National Science Foundation.

\bibliography{refs}

\clearpage
\appendix


\section{Discussion \& Examples of DPSGD Grammatical Errors}
\label{sec:appendix_dpsgd}

\paragraph{DPSGD repeats input instead of paraphrasing.}
In the paraphrasing task, we observed that models trained with DPSGD often produce output that is identical (or nearly identical) to the input. Non-private models and models trained with \ours occasionally reproduce portions of the input unchanged, but far less often than the DPSGD models---suggesting that DPSGD's noise interferes with the ability of the finetuning process to train models that effectively perform the desired task.


\paragraph{DPSGD misspells uncommon words.}
Both DPSGD and \ours models misspelled technical words (especially in the PubMed dataset) and names of people or places (especially in the XSum dataset). Misspellings seemed most common for uncommon words or names; for example, both models misspelled medical terms in the PubMed dataset (e.g. ``Parkinson's disease'' as ``Parkson's disease''), and DPSGD misspelled ``Hannah Ennis-Hill'' as ``Hillisis-hill'' in the XSum dataset, but both models correctly spelled common city names like ``Edinburgh.'' The \ours models produced many fewer misspellings than the DPSGD models, perhaps because the filtering step removed training examples with misspellings.

\paragraph{DPSGD makes grammatical errors and produces incomplete sentences.}
The DPSGD models made significant grammatical errors in all datasets. The DPSGD outputs often left out articles and mis-conjugated verbs; for example, in an example about flooding in Gujrat, the DPSGD model produced the sentence, ``death in floods in Gujarat have more than double in two days.'' \ours models had many fewer grammatical errors, and were comparable to non-private models, perhaps because the filtering process eliminates training examples with grammatical errors.

The DPSGD models also produced some sentences that were incomplete. For example, when summarizing an article about an election, the DPSGD model produced the sentence, ``out of the 28 candidates, 70 are women. the election will take place on 2 March.  the D are the party with the most candidates with 38  and the.'' \ours models produced complete sentences, perhaps because the entailment filtering step fails to infer the meaning of incomplete sentences.

\section{Phase-wise Dataset Generation and Models}
\label{appendix:phase_models}
Here we provide details on the models used across different phases of dataset generation for the MRPC, XSUM, and PubMed datasets (see Table \ref{tab:phase_models}. The substantial increase in the number of examples generated in Phase 3 highlights the effectiveness of self-distillation in enhancing the dataset and preparing it for task-specific fine-tuning.

\begin{table*}[h!]
\centering
\begin{tabular}{|c|p{2.5cm}|p{3cm}|p{2.2cm}|p{3cm}|p{2.5cm}|}

\hline
\textbf{Dataset} & \textbf{Phase 1 generation model} & \textbf{Phase 1 generated examples} & \textbf{Phase 2 fine-tuned model} & \textbf{Phase 3 generated examples} & \textbf{Phase 3 fine-tuned model} \\ \hline

MRPC  & GPT2   & 73706  & T5-large & 70143  & T5-large \\ \hline
XSUM  & GPT2   & 151779 & T5-large & 202707   & T5-large \\ \hline
Pubmed & BioGPT & 45484 & T5-large & 234048 & T5-large    \\ \hline
\end{tabular}
\caption{Phase-wise dataset generation and models used.}
\label{tab:phase_models}
\end{table*}

\section{Example AlpacalEval Prompt}
\label{sec:appendix_alpcal}

We show an example of the prompts used in AlpacaEval for evaluating language model performance based on fluency. This prompt is a modified version of the default AlpacaEval prompt with minor changes focused on the specific metric (fluency) for evaluation. All other prompts for the remaining six dimensions (e.g., coherence, consistency, relevance, etc.) were designed in a similar way to assess specific aspects of output quality.

\noindent\begin{minipage}{\linewidth}
\scriptsize
\begin{verbatim}
<|im_start|>system
You are a highly efficient assistant, who evaluates and selects 
the best large language model (LLMs) based on the quality of their 
responses to a given instruction. This process will be used to 
create a leaderboard reflecting the most accurate and 
human-preferred answers.
<|im_end|>
<|im_start|>user
I require a leaderboard for various large language models. 
I'll provide you with prompts given to these models and their 
corresponding outputs. 

Your task is to assess these responses, specifically for fluency, 
and select the model that demonstrates the highest fluency in 
its output.

## Instruction

{
    "instruction": """{instruction}""",
}

## Model Outputs

Here are the unordered outputs from the models. Each output is 
associated with a specific model, identified by a unique 
model identifier.

{
    {
        "model_identifier": "m",
        "output": """{output_1}"""
    },
    {
        "model_identifier": "M",
        "output": """{output_2}"""
    }
}

## Task

Evaluate the models based on how fluently they deliver their 
responses, and select the model that best demonstrates high 
fluency. Answer by providing the model identifier of the 
best model. We will use your output as the name of the 
best model, so make sure your output only contains one of 
the following model identifiers and nothing else 
(no quotes, no spaces, no new lines, ...): m or M.

## Best Model Identifier
<|im_end|>
\end{verbatim}
\end{minipage}
\normalsize

\section{Example Outputs}
\label{sec:appendix_hallucination}

\small

\subsection{XSum Dataset}

In this section, we present example outputs from the XSum dataset, showing the reference summaries and the summaries generated by non-private, DPSGD, and \(M_{refined}\) models. These examples highlight the differences in summarization quality across different models.

\begin{table*}[ht]
\small
\centering
\begin{tabular}{|l|p{12cm}|}
\hline
\rowcolor{mypink} \textbf{Reference Summary} & There is a ""chronic"" need for more housing for prison leavers in Wales, according to a charity. \\
\hline
\textbf{Non-private} &  A charity has said there is a ""desperate need"" for housing for ex-offenders after their release from prison in Wales. \\
\hline
\textbf{DPSGD} & prison leavers need help to find a home, charity says. the charity said there was   ""a desperate need for accommodation"" in Wales \\
\hline
\textbf{\(M_{refined}\)} & Police found three firearms, ammunition and a five-figure sum of money in Edinburgh's property on Wednesday. he was arrested and charged at Edinburgh Sheriff Court on Thursday." \\
\thickhline

\rowcolor{mypink} \textbf{Reference Summary} & Four people accused of kidnapping and torturing a mentally disabled man in a ""racially motivated"" attack streamed on Facebook have been denied bail. \\
\hline
\textbf{Non-private} &  Four teenagers accused of kidnapping and abducting a mentally ill man in Chicago have been denied bail, prosecutors say. \\
\hline
\textbf{DPSGD} & suspects charged with hate crimes and kidna of a man with disabilities  in  attack   man. sasts are accused of beating the victim, who suffers from schizophrenia and mental disorder \\
\hline
\textbf{\(M_{refined}\)} & The four were charged with hate crimes and aggravated kidnapping and battery in the case of a white man who was found dead in his home on Friday." \\
\thickhline

\rowcolor{mypink} \textbf{Reference Summary} & West Brom have appointed Nicky Hammond as technical director, ending his 20-year association with Reading. \\
\hline
\textbf{Non-private} & West Bromwich Albion have announced the retirement of former England goalkeeper Alan Shearer, who has been director of football at the club since 2003. \\
\hline
\textbf{DPSGD} & former Arsenal goalkeeper is West Brom's director of football. has been for the club since 2003. the former goal for for four years  at the Royals. \\
\hline
\textbf{\(M_{refined}\)} & West Brom said the former goalkeeper was a key player in the club's promotion to the Premier League in 2006 and 2012." \\
\thickhline

\rowcolor{mypink} \textbf{Reference Summary} & The pancreas can be triggered to regenerate itself through a type of fasting diet, say US researchers. \\
\hline
\textbf{Non-private} & A ""fasting-mimicking diet"" can help regenerate the pancreas, a study has shown, by starving mice for five days.. \\
\hline
\textbf{DPSGD} & scientists have found  diet can reverse diabetes in mice  by reprogramming cells in the pan. the diet is low in calories, high in fat and high fats \\
\hline
\textbf{\(M_{refined}\)} & Researchers have found that the diet can be used to reverse the effects of diabetes in mice by reprogramming the pancreas to respond to the sugar in the blood." \\
\thickhline

\rowcolor{mypink} \textbf{Reference Summary} & Since their impending merger was announced in January, there has been remarkably little comment about the huge proposed deal to combine Essilor and Luxottica.\\
\hline
\textbf{Non-private} & There is not a lot of interest in the proposed merger of Essilor and Luxottica, the two world's biggest eyewear companies. \\
\hline
\textbf{DPSGD} & silor and Luxottia are merging to create  the world's largest eyewear company Esilor. the deal is not  in the public interest, experts say ,  \\
\hline
\textbf{\(M_{refined}\)} & Essilor is the world's largest manufacturer of lenses and contact lenses, and Luxottica, the top of the frame industry, says the merger would be ""in the public interest"" but the industry is concerned that the deal could be a ""stanglehold"" on the supply of high-end glasses." \\
\thickhline

\rowcolor{mypink} \textbf{Reference Summary} & A ""medal at any cost"" approach created a ""culture of fear"" at British Cycling, says former rider Wendy Houvenaghel. \\
\hline
\textbf{Non-private} & British Cycling has apologised for ""failings"" in the organisation's World Class programme, according to former world champion Laura Houvenaghel.. \\
\hline
\textbf{DPSGD} & Olympic gold Honaghel British Cycling has admitted it has ""fas"" in the World Class programme, she says. she said the British had  ""zero regard"" for the athletes' welfare  \\
\hline
\textbf{\(M_{refined}\)} &Hounaghel said she was ""vindicated"" by the report into the treatment of athletes by British Cycling, which she said had ""sanitised"" the organisation's ""res"" after the Olympics." \\
\thickhline

\rowcolor{mypink} \textbf{Reference Summary} & Have you heard the one about the computer programmer who bought a failing comedy club in Texas and turned it into a million dollar a year business? \\
\hline
\textbf{Non-private} & ""We bought a sinking ship and we brought buckets to bail it out,"" says the owner of one of the US's best-known comedy clubs. \\
\hline
\textbf{DPSGD} & comedy clubs in the US are growing in popularity, says founder eem Badr. he owns the Hiout,  a comedy club  in Austin, Texas. the club was founded in 2009 after two friends bought it for \$20,000. \\
\hline
\textbf{\(M_{refined}\)} & ""The comedy industry is in its golden age,"" says Kareem Badr, who is the founder of the Hideout in Austin, Texas." \\
\thickhline

\rowcolor{mypink} \textbf{Reference Summary} & The reaction from BT's investors told us much about media regulator Ofcom's ruling on the fate of Openreach, the BT subsidiary that provides much of the UK's broadband infrastructure. \\
\hline
\textbf{Non-private} & Shareholders in BT have reacted with glee to the news that the government's competition watchdog, Ofcom, has said it will not take action against the company. \\
\hline
\textbf{DPSGD} & s  of  the  telecoms is to be split  by Ofcom. its  pension scheme, BT, has a deficit of £10b - \\
\hline
\textbf{\(M_{refined}\)} & BT's pension scheme, which is owned by the government, has been the subject of debate, and the Ofcom chief has said that the company is not going to be broken up." \\
\thickhline

\end{tabular}
\caption{Example summaries from non-private, DPSGD, and \(M_{refined}\) models on the XSum dataset.}
\label{table:xsum_summaries_clean2}
\end{table*}

\subsection{MRPC Dataset}

In this section, we present example paraphrases from the MRPC dataset, showing the reference paraphrases and the paraphrases generated by non-private, DPSGD, and \(M_{refined}\) models. These examples highlight the differences in paraphrasing quality across different models.

\begin{table*}[ht]
\small
\centering
\begin{tabular}{|l|p{12cm}|}
\hline
\rowcolor{mypink} \textbf{Reference Paraphrase} &PCCW 's chief operating officer , Mike Butcher , and Alex Arena , the chief financial officer , will report directly to Mr So . \\
\hline
\textbf{Non-private} & PCCW's board of directors will be led by chief executive John So, according to a statement from the organisation he heads. \\
\hline
\textbf{DPSGD} & PCCW's chief operating officer, Mike Butcher and chief financial officer Alex Arena will report directly to Mr So. Mr Arena is a member of the board of directors of pccw. \\
\hline
\textbf{\(M_{refined}\)} & PCCW's chief operating officer and chief financial officer will report to Mr So, who will be in charge of the company' s operations." \\
\thickhline

\rowcolor{mypink} \textbf{Reference Paraphrase} & The world 's two largest automakers said their U.S. sales declined more than predicted last month as a late summer sales frenzy caused more of an industry backlash than expected . \\
\hline
\textbf{Non-private} & Two of the world's largest automakers have said their U.S. sales fell more than expected last month as a late summer sales frenzy caused more of an industry backlash. \\
\hline
\textbf{DPSGD} & the world's two largest automakers said their U.S. sales declined more than predicted last month as a sales frenzy caused more of an industry backlash than expected.
 \\
\hline
\textbf{\(M_{refined}\)} & The world's two largest automakers said their U.S. sales fell more than expected last month as a late summer sales surge slowed sales." \\
\thickhline

\rowcolor{mypink} \textbf{Reference Paraphrase} & According to the federal Centers for Disease Control and Prevention ( news - web sites ) , there were 19 reported cases of measles in the United States in 2002 . \\
\hline
\textbf{Non-private} & The measles is a highly contagious disease that can be fatal if it is not treated in time, according to the World Health Organization ( WHO ). \\
\hline
\textbf{DPSGD} & According to the federal Centers for Disease Control and Prevention ( news - web sites ), there were 19 reported cases of measles in the United States in 2002. \\
\hline
\textbf{\(M_{refined}\)} & In 2002, the U.S. Centers for Disease Control and Prevention reported 19 cases of measles in the United States." \\
\thickhline

\rowcolor{mypink} \textbf{Reference Paraphrase} & A tropical storm rapidly developed in the Gulf of Mexico Sunday and was expected to hit somewhere along the Texas or Louisiana coasts by Monday night .\\
\hline
\textbf{Non-private} & A tropical storm is expected to make landfall in the Gulf of Mexico on Monday night, the National Hurricane Center (NHC) has said. \\
\hline
\textbf{DPSGD} & A tropical storm rapidly developed in the Gulf of Mexico Sunday and was expected to hit somewhere along the Texas or Louisiana coasts by Monday night. \\
\hline
\textbf{\(M_{refined}\)} & A tropical storm was developing in the Gulf of Mexico Sunday and was expected to make landfall in Texas or Louisiana by Monday night, according to the National Weather Service." \\
\thickhline

\rowcolor{mypink} \textbf{Reference Paraphrase} & The company didn 't detail the costs of the replacement and repairs . \\
\hline
\textbf{Non-private} & The company has said it will replace the car's engine, but hasn't said how much it would cost to replace it with a new one.\\
\hline
\textbf{DPSGD} & The company didn 't detail the costs of the replacement and repairs. the company did not provide a breakdown of costs. \\
\hline
\textbf{\(M_{refined}\)} & The company did not provide a cost estimate for the replacements or repairs. - paraphrased , pg." \\
\thickhline

\rowcolor{mypink} \textbf{Reference Paraphrase} & The settling companies would also assign their possible claims against the underwriters to the investor plaintiffs , he added . \\
\hline
\textbf{Non-private} & The settling companies would also assign their possible claims against the underwriters to the investor plaintiffs, he said ; paraphrase. \\
\hline
\textbf{DPSGD} & The settling companies would also assign their possible claims against the underwriters, the investor plaintiffs added. ' \\
\hline
\textbf{\(M_{refined}\)} & The settling companies would also assign the investor plaintiffs' claims against the underwriters, he added, noting that the companies had not yet settled." \\
\thickhline

\rowcolor{mypink} \textbf{Reference Paraphrase} & Air Commodore Quaife said the Hornets remained on three-minute alert throughout the operation . \\
\hline
\textbf{Non-private} & The Royal Navy's Hornets remained on high alert throughout the operation, the Royal Air Force has said, paraphrasing. \\
\hline
\textbf{DPSGD} & Air Commodore Quaife said the Hornets remained on three-minute alert throughout the operation. \\
\hline
\textbf{\(M_{refined}\)} & The Hornets were on three-minute alert during the operation, Air Commodore Quaife said. - paraphrased" \\
\thickhline

\rowcolor{mypink} \textbf{Reference Paraphrase} &  A Washington County man may have the countys first human case of West Nile virus , the health department said Friday . \\
\hline
\textbf{Non-private} & A Washington County man may have the first human case of West Nile virus, the health department has said. Paraphrased \\
\hline
\textbf{DPSGD} &  A Washington County man may have the countys first human case of West Nile virus, the health department said Friday. \\
\hline
\textbf{\(M_{refined}\)} & The state health department said Friday that a man in Washington County may have the countys first West Nile virus case, the first in the state." \\
\thickhline

\rowcolor{mypink} \textbf{Reference Paraphrase} & Moseley and a senior aide delivered their summary assessments to about 300 American and allied military officers on Thursday . \\
\hline
\textbf{Non-private} & US Secretary of Defense Robert Moseley and a senior aide have delivered their summary assessments of the Iraq war to senior military officers in Washington. \\
\hline
\textbf{DPSGD} & Moseley and a senior aide delivered their summary assessments to about 300 allied military officers on Thursday. he paraphrase: \\
\hline
\textbf{\(M_{refined}\)} & Moseley and a senior adviser delivered the summary assessments to about 300 U. S. and NATO military officers on Thursday. - paraphrased.". \\
\thickhline

\end{tabular}
\caption{Example paraphrases from non-private, DPSGD, and \(M_{refined}\) models on the MRPC dataset.}
\label{table:mrpc_paraphrases_clean2}
\end{table*}

\begin{table*}[ht]
\small
\centering
\begin{tabular}{|l|p{12cm}|}
\hline
\hline
\rowcolor{mypink} \textbf{Reference Summary} &  --- \\
\hline
\textbf{Non-private} & background : congenital adrenal hyperplasia ( cah ) refers to a group of autosomal recessive disorders caused by an enzyme deficiency which leads to defects in biosynthesis of steroid precursors. during childhood, the management is largely focused on achieving normal growth and attaining appropriate final adult height.aim:the aim of this study was to evaluate the clinical and biochemical features of patients diagnosed to have \\
\hline
\textbf{DPSGD} & the relationship between anxiety and cogn in  parkson's disease (pd) has not been explored.   this study investigated whether anxiety differentially affects cogn  in patients with left-sided dominant d (ldd ) compared p patients  with anxiety. the results showed that apathy, but not anxiety, was associated with executive dysfunction,  and pathy. in the lp  patients,pathy was not associated  cognitive 
 \\
\hline
\textbf{\(M_{refined}\)} & anxiety affects quality of life in those living with parkinson's disease (pd) more than overall cognitive status, motor deficits, apathy, and depression, but the relationship between anxiety and cognition in d is still meager and lags far behind that of depression. however, recent studies have shown that anxiety is more prominent and prevalent than depression and is associated with cognitive performance."\\
\thickhline

\rowcolor{mypink} \textbf{Reference Summary} & --- \\
\hline
\textbf{Non-private} & mirnas are a class of multifunctional singled - stranded small RNAs which regulate gene expression and play important roles in the onset and progression of tumorigenesis. the current treatment of radiation therapy, chemotherapy and hormone therapy slow mammary gland tumor growth and prolong survival but metastatic breast cancer still remains incurable due to our limited understanding of the molecular mechanisms that drive tumor formation and metastasis in eukary \\
\hline
\textbf{DPSGD} & nns are non-coding rs tran  into mna but remain untranslated in e otic cells. they includern, mira,pirn andsnolarrra. mirn are  class of multi  d-stranded small aa 20nt in length and regulate  range of processes including development, cell cycle progression, stem cell renewal, differentiation, proliferation, metsis and \\
\hline
\textbf{\(M_{refined}\)} & rnas are transcribed into mra but remain untranslated in eukaryotic cells. they include sir ( small interfering ), mir ( micror) and snolar (small nucleolar ). mira is   class of single - stranded small RNAs that regulate expression and expression of genes and are involved in the initiation and progression of tumorigenesis." \\
\thickhline

\rowcolor{mypink} \textbf{Reference Summary} & --- \\
\hline
\textbf{Non-private} & ovarian hyperstimulation syndrome ( ohss ) is a serious complication of in vitro fertilization, occurring in 1 - 10\% of patients. vascular endothelial growth factor has emerged as one of the factors most likely involved in the pathophysiology of this iatrogenic condition and human chorionic gonadotrophin has been implicated in its etiopathogenesis  \\
\hline
\textbf{DPSGD} & hs is  complication of oulation induction, occurring in 1- 10\% of ino fertilization patients. it has spectrum of clinical and laboratory manifestationsranging from mild to life-threatening conditions.vascular endhelial growth factor has emerged as one factor most likely involved in the pathophysio ofos.  endoheli growth is an ogenic kin that is stimulator of thevascular thelium with its sequelae of ascite and \\
\hline
\textbf{\(M_{refined}\)} & hs is a serious complication of ovulation induction, occurring in 1 - 10\% of in vitro fertilization patients. vascular endothelial growth factor is an angiogenic factor that causes capillary permeability with its sequelae of ascites and pleural effusion. the early form is caused by exogenous chorionic gonadotrophin (hcg) and occurs before the 10" \\
\thickhline

\rowcolor{mypink} \textbf{Reference Summary} & --- \\
\hline
\textbf{Non-private} & background : congenital adrenal hyperplasia ( cah ) refers to a group of autosomal recessive disorders caused by an enzyme deficiency which leads to defects in biosynthesis of steroid precursors. during childhood, the management is largely focused on achieving normal growth and attaining appropriate final adult height.aim:the aim of this study was to evaluate the clinical and biochemical features of patients diagnosed to have \\
\hline
\textbf{DPSGD} & congenital hyperpa refers to  group of autosom reces disorders caused by an enzyme de which leads to defects in biosynthesis ofsteroid precursors. the non classical cah variant is more common with prevalence of 1 in 1000   the classical h is  more  common  with a prevalence rate of 1000.  management is focused on normal growth and attaining appropriate adult height and  height sds andbi  data were collected \\
\hline
\textbf{\(M_{refined}\)} & congenital adrenal hyperplasia ( cah) is   group of autosomal recessive disorders caused by an enzyme deficiency which leads to defects in synthesis of steroid precursors. johns medical college hospital, bangalore by the department of endocrinology on patients diagnosed to have ca and seen in outpatient clinic between january 2012 and october 2012. all patients underwent bio" \\
\thickhline

\end{tabular}
\caption{Example summaries from non-private, DPSGD, and \(M_{refined}\) models on the PubMed dataset.}
\label{table:pubmed_paraphrases_clean}
\end{table*}

\begin{table*}[ht]
\small
\centering
\begin{tabular}{|l|p{12cm}|}
\hline
\hline
\rowcolor{mypink} \textbf{Reference Summary} & ---- \\
\hline
\textbf{Non-private} & type 1 diabetes ( t1d ) results from the destruction of insulin producing pancreatic cells by a cell specific autoimmune process. in order to obtain insight into the pathogenic mechanisms in humans and to test novel therapeutic approaches for its treatment, different preclinical models of the disease such as spontaneous and accelerated diabetes in the non - obese diabetic ( nod) mice or diabetes induced in susceptible rodent strains by multiple low doses of str \\
\hline
\textbf{DPSGD} & pentoxlline (px) is a xanthin-derived general phosphosterase ( (de) inhibitor that has immunomodulatory and anti-inflammatory activity. tx is an  p -  inhibitor of dieas  ( de ), which has been available for many years to treat vascular disorder. the st  and cc phosphate (cgp) andcic \\
\hline
\textbf{\(M_{refined}\)} & pt is  a methyl xanthine-derived general phosphodiesterase (pde) inhibitor that has been used for many years to treat vascular disorders, and it is thought to have anti- inflammatory activity, which may be useful in the treatment of type 1 diabetes. t1d is caused by the destruction of insulin producing pancreatic cells by  specific autoimmune process. the pathogenesis of  1d was mediated by" \\
\thickhline

\rowcolor{mypink} \textbf{Reference Summary} & --- \\
\hline
\textbf{Non-private} & objective : to determine the presence of specific gastrointestinal antimicrobials and their sensitivity, as well as whether these are associated with severity of asthma in adults.methods:a cross - sectional study including adult asthmatics in ambulatory care at a tertiary care hospital in the city of rio de janeiro. patients were randomly assigned to two groups of severity according to the global initiative for asthma criterion  \\
\hline
\textbf{DPSGD} & os  srics de antipos ige antitas estoccicas, as  asa em adulto  in hospital universitrio no rii (rj)  study included adultms in acomament ambulatorial  hospital universityir io dejano ( j.  determined nves specifics of corposige andoxinsaafil \\
\hline
\textbf{\(M_{refined}\)} & estudo transversal incluindo asmticos adulto  ambulatorial  in um hospital universitrio no janeiro (rj). o pacientes foram alocados consecutivamente in two groups de gravidade da asa segundo criteria of global initiative for asthma: as leve ( al ) and asmoderada ou grave ( amg ). \\
\thickhline

\rowcolor{mypink} \textbf{Reference Summary} & --- \\
\hline
\textbf{Non-private} & background : the family is the cornerstone of human social support network and its presence is essential in everyone s life. changes inevitably occur in families with illness and hospitalization of a family member.objectives:the purpose of this study was to investigate the satisfaction of the families of patients in the intensive care unit ( icu ) in tehran.methods:this cross - sectional survey was conducted on the patients, their families and their
  \\
\hline
\textbf{DPSGD} & the satisfaction of the family of hospitalized patients in intensive care units was investigated. the families of intensive units are facing life-threatening illness experiences and complex treatments, different technologies and different equipment are used  which could lead to the dissatis of family members. this study aimed to investigate satisfaction satisfaction  of families in the intensive unit. results showed that satisfaction family satisfaction was higher in  intensive  units than in other , the  satisfaction and satisfaction were higher  than the \\
\hline
\textbf{\(M_{refined}\)} & the family is the cornerstone of human social network and its presence is essential in everyone s life, and changes inevitably occur in families with illness and hospitalization, especially intensive care unit (icu). the needs of the families of hospitalized patients in the i are summarized in five sections, which include: assurance, communication, information, support, comfort and satisfaction. fox et al. found that the satisfaction of family members is   the most important" \\
\thickhline

\rowcolor{mypink} \textbf{Reference Summary} & --- \\
\hline
\textbf{Non-private} & Non-private: background. coronary artery disease is one of the most widespread chronic diseases in most countries and it is associated with high mortality, limitations in life and disability in a large part of productive forces of iran.objectives.the aim of this study was to investigate the relationship between anxiety and depression in patients undergoing heart surgery.patients and methods.a cross - sectional survey was conducted among patients who had undergone cardiac surgery from january 2011 to \\
\hline
\textbf{DPSGD} & coronaryartery disease is the most cardiovascular disorder in developing and developed countries. it is not only   chronic disease with high mortality, but it causes limitations in life and disability in a large part of productive forces of the country.  it also associated with increased production and medical costs. coronar artery bypass surgery is one of surgical procedures for treatment of cardiovascular patients. on the other, complications are anxiety and depression. anxiety is  common psychological response of patients which is associated quality of \\
\hline
\textbf{\(M_{refined}\)} & the development of human societies and industrialization as well as changes in stress sources has changed the disease pattern in civilized societies. coronary artery disease is the most widespread chronic disease in most countries. according to forecasts, the mortality rate in eastern mediterranean countries including iran will be 30 - 35\%. anxiety and depression are associated with the quality of life of patients with cardiovascular diseases. in fact, anxiety is a common psychological response of cardiovascular patients."\\
\thickhline

\rowcolor{mypink} \textbf{Reference Summary} & --- \\
\hline
\textbf{Non-private} & objective : influenza a virus ( h3-f7 ) was isolated from patients in nepal, thailand and the united states in 2009. rna was extracted from 48-hour shell vial cultures and analyzed by reverse transcription - polymerase chain reaction and ferret antisera using the superscript pcr system ( invitrogen ; carlsbad / california  \\
\hline
\textbf{DPSGD} &ra was extracted from 48-hour shell cultures by using the magnapur lx (roche mocular) andrii kiti ( roch mo )   according to the manufacturer's protocols. t-pr amplification was performed by the one- step reverse - polymerase reaction system (rogen, carbad, ca, usa). gtg- ct  atc-gc \\
\hline
\textbf{\(M_{refined}\)} & rna was extracted from 48-hour shell cultures ( 10 ) and was amplification by reverse transcription- polymerase chain reaction (rt-pcr). the swab samples were collected within the first 72 hours of onset of symptoms, and were analyzed by immunoassay and monoclonal antibody staining ( 11 ). for antigenic analysis of select isolates, mra  was added to" \\
\thickhline
\end{tabular}
\caption{Example summaries from non-private, DPSGD, and \(M_{refined}\) models on the PubMed dataset.}
\label{table:pubmed_paraphrases_clean2}
\end{table*}

\begin{table*}[ht]
\small
\centering
\begin{tabular}{|l|p{12cm}|}
\hline
\rowcolor{mycolor} \textbf{Original Text} & PCCW's chief operating officer, Mike Butcher, and Alex Arena, the chief financial officer, will report directly to Mr So. \\
\hline
\textbf{\(M_{base}\)} & The company's chief operating officer, Mike Butcher, and chief financial officer Alex Arena will report directly to Mr So and will not be subject to the oversight of the PCCW board of directors. \\
\hline
\textbf{\(M_{private}\)} & The company’s chief operating officer, Mike Butcher, and chief financial officer Alex Arena will report directly to Mr So, who will be in charge of the company’s operations. \\
\hline
\textbf{\(M_{refined}\)} & PCCW's chief operating officer and chief financial officer will report to Mr So, who will be in charge of the company’s operations. \\
\hline
\thickhline

\rowcolor{mycolor} \textbf{Original Text} & The world’s two largest automakers said their U.S. sales declined more than predicted last month as a late summer sales frenzy caused more of an industry backlash than expected. \\
\hline
\textbf{\(M_{base}\)} & The world’s two largest automakers, General Motors and Chrysler, said their U.S. sales declined more than they had expected last month, as a weak economy and weak consumer demand weighed on their bottom lines. \\
\hline
\textbf{\(M_{private}\)} & The world’s two largest automakers, Ford and General Motors, said their U.S. sales declined more than expected last month as a slump in sales slowed sales of both cars and trucks. \\
\hline
\textbf{\(M_{refined}\)} & The world’s two largest automakers said their U.S. sales fell more than expected last month as a late summer sales surge slowed sales. \\
\hline
\thickhline

\rowcolor{mycolor} \textbf{Original Text} & According to the federal Centers for Disease Control and Prevention (CDC), there were 19 reported cases of measles in the United States in 2002. \\
\hline
\textbf{\(M_{base}\)} & According to the Centers for Disease Control and Prevention (CDC), there were 19 cases of measles in the United States in 2002. \\
\hline
\textbf{\(M_{private}\)} & According to the Centers for Disease Control and Prevention (CDC), there were 19 cases of measles in the United States in 2002. \\
\hline
\textbf{\(M_{refined}\)} & In 2002, the U.S. Centers for Disease Control and Prevention reported 19 cases of measles in the United States. \\
\hline
\end{tabular}
\caption{Example paraphrases from \(M_{base}\), \(M_{private}\), and \(M_{refined}\) models on the MRPC dataset.}
\label{table:mrpc_paraphrases_clean}
\end{table*}

\section{Example Outputs for Different Phases}
\label{phase_output}
In this section, we present example outputs from the XSum and MRPC dataset, showing the reference summaries/paraphrases and the summaries/paraphrases generated by \(M_{base}\), \(M_{private}\), and \(M_{refined}\) models. These examples highlight the differences in summarization quality across the different stages of DPRefine.

\begin{table*}[ht]
\small
\centering
\begin{tabular}{|l|p{12cm}|}
\hline
\rowcolor{mycolor} \textbf{Reference Summary} & There is a "chronic" need for more housing for prison leavers in Wales, according to a charity. \\
\hline
\textbf{\(M_{base}\)} & Prison Link Cymru said it had received 1,099 referrals in 2015-16, with some ex-offenders living on the streets for up to a year before being able to find accommodation. \\
\hline
\textbf{\(M_{private}\)} & Prisons in Wales have been more than ever, but the charity says they are struggling to find accommodation for offenders in the country. \\
\hline
\textbf{\(M_{refined}\)} & Wales' prisons have said that the number of people living on the streets is increasing, but the charity says the need for more one-bedroom flats is more than ever. \\
\hline
\thickhline

\rowcolor{mycolor} \textbf{Reference Summary} & A man has appeared in court after firearms, ammunition and cash were seized by police in Edinburgh. \\
\hline
\textbf{\(M_{base}\)} & Officers searched properties in the city’s Waterfront Park and Colonsay View on Wednesday. A 26-year-old man was arrested and charged with robbery. \\
\hline
\textbf{\(M_{private}\)} & Police found three firearms and a five-figure sum of money in Edinburgh on Wednesday. A man was arrested and charged in the attack in the city. \\
\hline
\textbf{\(M_{refined}\)} & Police found three firearms, ammunition, and a five-figure sum of money in Edinburgh’s property on Wednesday. He was arrested and charged at Edinburgh Sheriff Court on Thursday. \\
\hline
\thickhline


\rowcolor{mycolor} \textbf{Reference Summary} & Four people accused of kidnapping and torturing a mentally disabled man in a "racially motivated" attack streamed on Facebook have been denied bail. \\
\hline
\textbf{\(M_{base}\)} & Prosecutors said the suspects stuffed a sock into the victim’s mouth, taped his mouth shut, and beat the 18-year-old white victim to death, according to the Chicago Tribune. \\
\hline
\textbf{\(M_{private}\)} & Suspects are accused of beating the 18-year-old victim in a van after he was taken to the police station, police said on Friday. \\
\hline
\textbf{\(M_{refined}\)} & The four were charged with hate crimes and aggravated kidnapping and battery in the case of a white man who was found dead in his home on Friday. \\
\hline
\end{tabular}
\caption{Example summaries from \(M_{base}\), \(M_{private}\), and \(M_{refined}\) models on the XSum dataset.}
\label{table:xsum_summaries_clean}
\end{table*}

\end{document}